\documentclass[10pt,journal,compsoc]{IEEEtran}

\usepackage{times}
\usepackage{graphics}
\usepackage{graphicx}
\usepackage{amsmath}
\usepackage{amsthm}
\usepackage{amssymb}
\usepackage{algorithm,algpseudocode}

\usepackage{tabularx}
\usepackage{multirow}
\usepackage{makecell}
\usepackage{booktabs}
\usepackage{overpic}
\usepackage{color}

\ifCLASSOPTIONcompsoc
  \usepackage[nocompress]{cite}
\else
  \usepackage{cite}
\fi

\usepackage[pagebackref=false,breaklinks=true,letterpaper=true,colorlinks,bookmarks=false]{hyperref}
\newif\ifdraft
\draftfalse
\drafttrue

\definecolor{orange}{rgb}{1,0.5,0}
\definecolor{violet}{RGB}{70,0,170}
\definecolor{pink}{RGB}{252,107,252}
\definecolor{brown}{RGB}{139,69,19}
\definecolor{red_fig}{RGB}{189,9,9}

 \newcommand{\ER}[1]{{\color{violet}{\bf ER: #1}}}
 \newcommand{\er}[1]{{\color{violet} #1}}
 \newcommand{\AL}[1]{{\color{blue}{\bf AL: #1}}}
 
 \newcommand{\SR}[1]{{\color{green}{\bf SR: #1}}}

\newcommand{\comment}[1]{}

\newcommand{\bc}{\mathbf{c}}
\newcommand{\bn}{\mathbf{n}}
\newcommand{\bp}{\mathbf{p}}

\newcommand{\bu}{\mathbf{u}}
\newcommand{\bv}{\mathbf{v}}
\newcommand{\bz}{\mathbf{z}}
\newcommand{\bx}{\mathbf{x}}

\newcommand{\myfootnotesize}{\fontsize{8pt}{10pt}\selectfont}
\newtheorem{theorem}{Theorem}

\newcommand{\dmesh}{\textit{DeepMesh}}

\begin{document}
\title{DeepMesh: Differentiable Iso-Surface Extraction}

\author{Beno\^it Guillard, Edoardo Remelli, Artem Lukoianov, Stephan R. Richter, Timur Bagautdinov, Pierre Baque, Pascal~Fua,~\IEEEmembership{Fellow,~IEEE}%
	\IEEEcompsocitemizethanks{
		\IEEEcompsocthanksitem B. Guillard , E. Remelli and P. Fua are with the Computer Vision Laboratory, \'{E}cole Polytechnique F\'{e}d\'{e}rale de Lausanne, Switzerland. E-mail: \{benoit.guillard, edoardo.remelli, pascal.fua\}@epfl.ch. A. Lukoianov is with the Group of Geometrical Data Processing at MIT. E-mail: arteml@mit.edu, P. Baque is with Neural Concept. E-mail: pierre.baque@neuralconcept.com. S.R. Richter is with Interl Labs.. T. Bagautdinov is with Meta Reality Labs. E-mail: timurb@meta.com. }
}

\markboth{ }
		{Shell \MakeLowercase{\textit{et al.}}: Bare Demo of IEEEtran.cls for Computer Society Journals}

\IEEEtitleabstractindextext{

\begin{abstract}
Geometric Deep Learning has recently made striking progress with the advent of continuous deep implicit fields. They allow for detailed modeling of watertight surfaces of arbitrary topology while not relying on a 3D Euclidean grid, resulting in a learnable parameterization that is unlimited in resolution. 
Unfortunately, these methods are often unsuitable for applications that require an \textit{explicit} mesh-based surface representation because converting an implicit field to such a representation relies on the Marching Cubes algorithm, which cannot be differentiated with respect to the underlying implicit field.
In this work, we remove this limitation and introduce a differentiable way to produce explicit surface mesh representations from Deep Implicit Fields. Our key insight is that by reasoning on how implicit field perturbations impact local surface geometry, one can ultimately differentiate the 3D location of surface samples  with respect to the underlying deep implicit field. We exploit this to define \textit{DeepMesh} --- an end-to-end differentiable mesh representation that can vary its topology.
We validate our theoretical insight through several applications: Single view 3D Reconstruction via Differentiable Rendering, Physically-Driven Shape Optimization, Full Scene 3D Reconstruction from Scans and End-to-End Training. In all cases our end-to-end differentiable parameterization gives us an edge over state-of-the-art algorithms.
\end{abstract}

}

\maketitle

\IEEEdisplaynontitleabstractindextext

\IEEEpeerreviewmaketitle


\section{Introduction}

Geometric Deep Learning has recently witnessed a breakthrough with the advent of Deep Implicit Fields (DIFs) \cite{Park19c,Mescheder19,Chen19c}. These enable detailed modeling of watertight surfaces without relying on a 3D Euclidean grid or meshes with fixed topology, resulting in a learnable surface parameterization that is \textit{not} limited in resolution.

However, a number of important applications require {\it explicit} surface representations, such as triangulated meshes or 3D point clouds. Computational Fluid Dynamics (CFD) simulations and the associated learning-based surrogate methods used for shape design in many engineering fields~\cite{Baque18,Umetani18} are a good example of this where 3D meshes serve as boundary conditions for the Navier-Stokes equations. Similarly, many advanced physically-based rendering engines require surface meshes to model the complex interactions of light and physical surfaces efficiently~\cite{Nimier19,Pharr16}.

Making explicit representations benefit from the power of deep implicit fields requires converting the implicit surface parameterization to an explicit one, which typically relies on one of the many variants of the Marching Cubes algorithm~\cite{Lorensen87,Newman06}. However, these approaches are not fully differentiable~\cite{Liao18a}. This makes it difficult to use continuous deep implicit fields to parameterize explicit surface meshes. 


\begin{figure*}[t]  
		\begin{center}
			\begin{overpic}[clip, trim=0.0cm 10cm 0 0.5cm,width=1.0\textwidth]{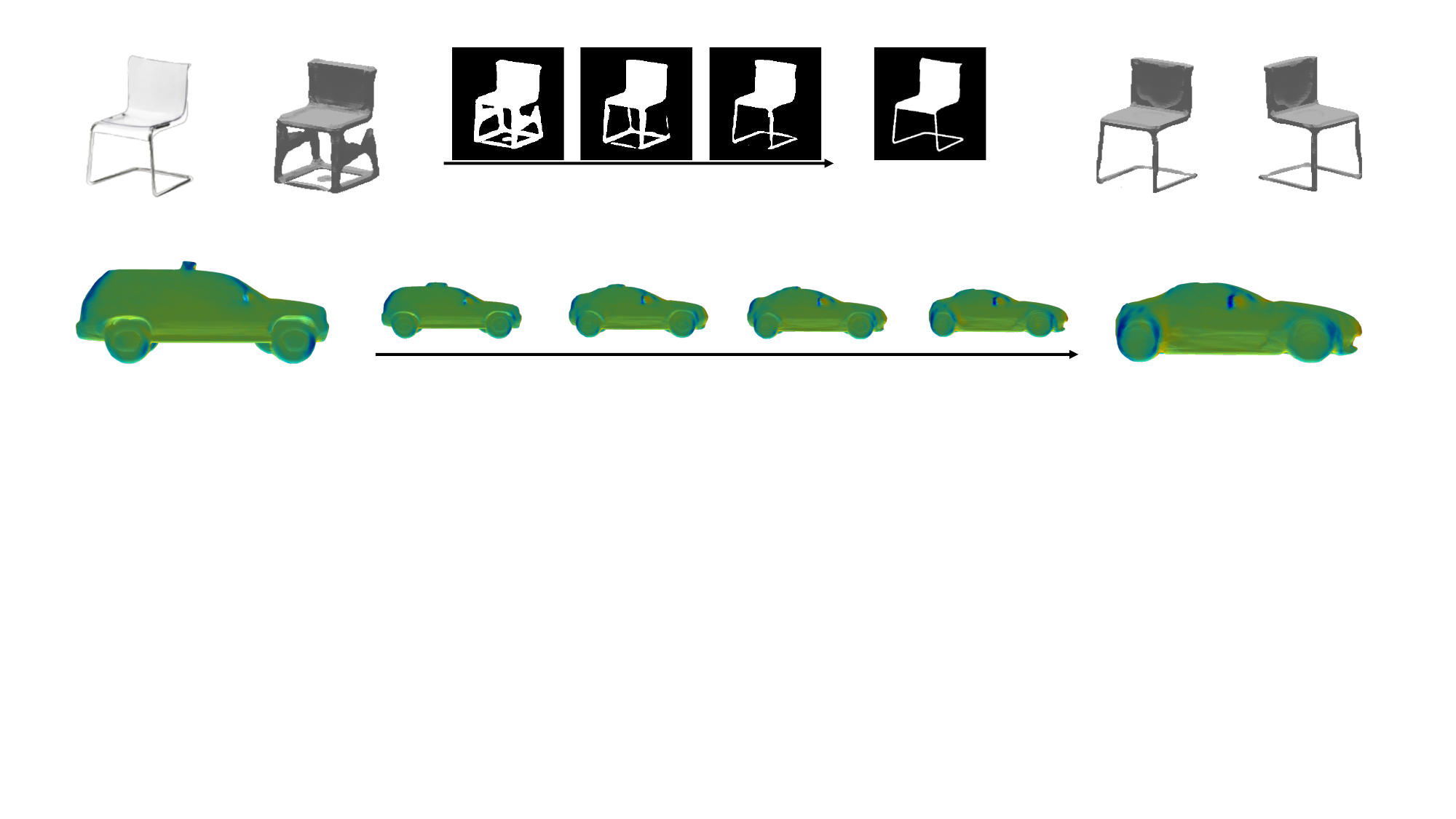}
				\put(6,10.5){\small{Image }}
				\put(16,10.5){\small{\textit{DeepMesh} Raw}}
				\put(35,12.5){\small{Silhouette refinement}}
				\put(58.5,12.5){\small{Target silhouette}}
				\put(76,10.5){\small{\textit{DeepMesh} Refined}}
				\put(6,0.0){\small{\textit{DeepMesh} Raw}}
				\put(40,0.0){\small{Drag minimization}}
				\put(76,0.0){\small{\textit{DeepMesh} Refined}}
				\put(0,18.5){\small{(a) }}
				\put(0,4.5){\small{(b) }}
			\end{overpic}
			\begin{overpic}[clip, trim=0.0cm 0cm 0 -1cm,width=1.0\textwidth]{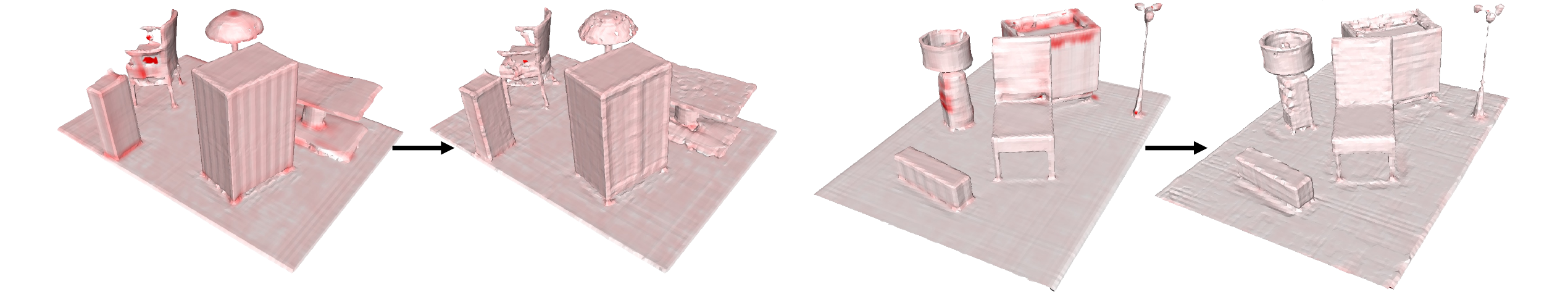}
				\put(0,10){\small{(c) }}
				\put(12,-1.0){\small{Raw}}
				\put(23,7){\small{Point cloud}}
				\put(25,5.5){\small{fitting}}
				\put(34,-1.0){\small{Refined}}

				\put(60,-1.0){\small{Raw}}
				\put(70,7){\small{Point cloud}}
				\put(72,5.5){\small{fitting}}
				\put(80,-1.0){\small{Refined}}
			\end{overpic}
		\end{center}
		\vspace{0pt}
		\caption{\textbf{DeepMesh}. (a) We condition our representation on an input image and output an initial 3D mesh, which we refine via differentiable rasterization \cite{Kato18}, thereby exploiting DeepMesh's end-to-end differentiability. (b) We use our parameterization as a powerful regularizer for aerodynamic optimization tasks. Here, we start from an initial car shape and refine it to minimize pressure drag. (c) We use the end-to-end differentiability of iso-surface extraction to improve the occupancy field fitted to a sparse point cloud of a whole scene by an off-the-shelf network, {Convolutional Occupancy Network (\textit{CON})~\cite{Peng20c}}. In these two examples, the raw output of~\cite{Peng20c} is shown on the left and the refined version on the right. The errors are shown in red and are smaller after refinement.}  
\label{fig:teaser}
\vspace{-6pt}
\end{figure*}
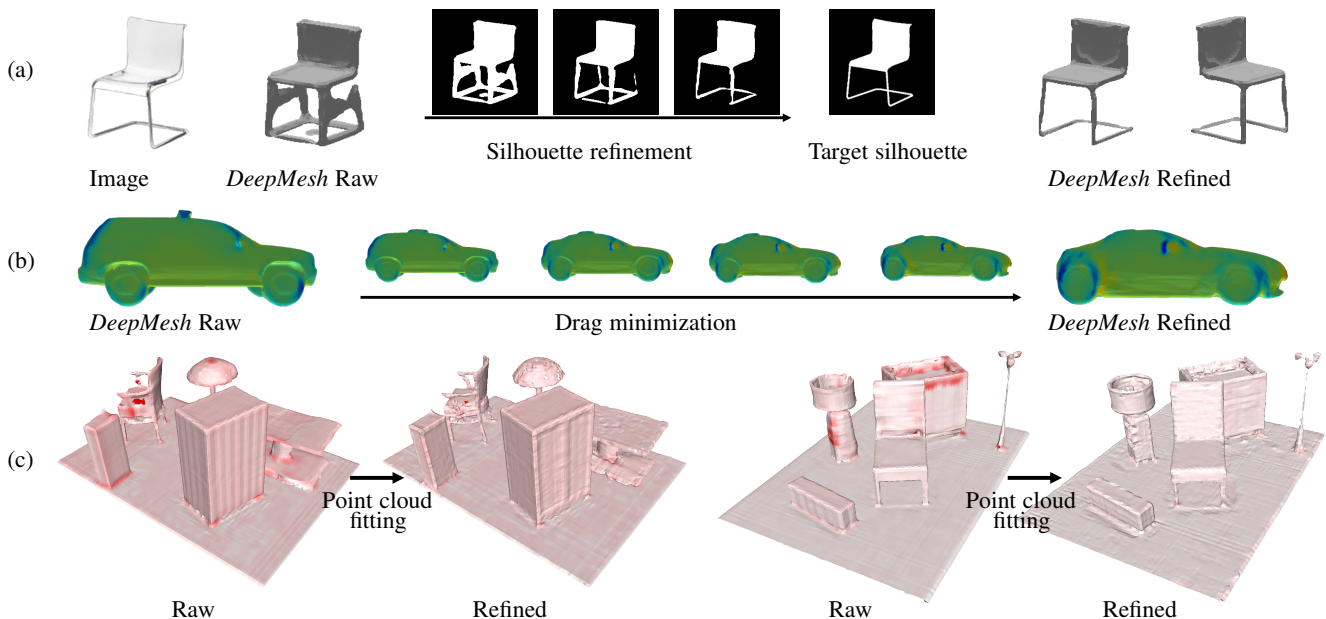

The non-differentiability of Marching Cubes has been addressed by learning differentiable approximations of it~\cite{Liao18a,Wickramasinghe20}. These techniques, however, remain limited to low-resolution meshes~\cite{Liao18a} or fixed topologies~\cite{Wickramasinghe20}. An alternative approach is to reformulate downstream tasks, such as differentiable rendering~\cite{Jiang20a,Liu19j} or surface reconstruction~\cite{Michalkiewicz19}, directly in terms of implicit functions, so that explicit surface representations are no longer needed. However, doing so is not easy and may even not be possible for more complex tasks, such as solving CFD optimization problems.

By contrast, we show that it is possible to use implicit functions, be they signed distance functions or occupancy maps, to produce explicit surface representations while preserving differentiability.  Our key insight is that 3D surface samples {\it can} be differentiated with respect to the underlying deep implicit field. We prove this formally by reasoning about how implicit field perturbations impact 3D surface geometry \textit{locally}. Specifically, we derive a closed-form expression for the derivative of a surface sample with respect to the underlying implicit field, which is independent of the method used to compute the isosurface.  
This lets us extract the explicit surface using a non-differentiable algorithm, such as Marching Cubes, and then perform the backward pass through the extracted surface samples. This yields an end-to-end differentiable surface parameterization that can describe arbitrary topology and is not limited in resolution. We will refer to our approach as \dmesh{}. We first introduced it in a conference paper~\cite{Remelli20b} that focused on {the $0$-isosurface} of signed distance functions. We extend it here to isosurface of generic implicit functions, such as occupancy fields by harnessing simple multivariate calculus tools.

We showcase the power and versatility of \dmesh{} in several applications. 
\begin{enumerate}

 \item Given a model trained to map latent vectors to SDFs, we use our approach to triangulate the SDF fields and write image-based losses that yield improved 3D reconstructions from single images, as shown in Fig.~\ref{fig:teaser}(a).  
 
 \item Similarly, we use the surface triangulations to compute the aerodynamic properties of 3D shapes and refine them, as shown in Fig.~\ref{fig:teaser}(b).  
 
 \item  We use our paradigm in conjunction with DIF-based methods to improve their performance in a  plug-and-play fashion by adding loss terms that can be computed on the meshes. This highlights the importance to be able to handle both SDFs and occupancy grids.
 
 \item We demonstrate that we can use our approach not only to better exploit the results of pre-trained networks but to actually train them better.

\end{enumerate}

In all these cases, our end-to-end differentiable parameterization gives us an edge over state-of-the art algorithms. In short, our core contribution is a theoretically well-grounded and computationally efficient way to differentiate through iso-surface extraction. This enables us to harness the full power of neural implicit fields to define an end-to-end differentiable surface mesh parameterization that allows topology changes.

 

\section{Related Work}
\label{sec:related}

\subsection{From Discrete to Continuous Implicit Surfaces.}

Level sets of a 3D function can represent watertight surfaces whose topology  can change~\cite{Sethian99,Osher03}. Being representable as 3D grids and thus easily processable by standard deep learning architectures, they have been used extensively~\cite{Brock16,Choy16b,Gadelha16,Rezende16,Riegler17,Tatarchenko17,Wu15b,Xie19a}. However, methods operating on dense grids have been limited to low resolution volumes due to excessive memory requirements. Methods operating on sparse representations of the grid tend to trade off the need for memory for a limited representation of fine details and lack of generalization~\cite{Richter18,Riegler17,Tatarchenko17,Tatarchenko19}.

This has changed recently with the introduction of continuous deep implicit fields, which represent 3D shapes as level sets of deep networks that map 3D coordinates to a signed distance function~\cite{Park19c} or occupancy field~\cite{Mescheder19,Chen19c}. This yields a continuous shape representation with respect to 3D coordinates that is lightweight but not limited in resolution. This representation has been successfully used for single view 3D reconstruction~\cite{Mescheder19,Chen19c,Xu19b}, and 3D shape completion~\cite{Chibane20a, Peng20}. 

{Signed distance and occupancy fields have case-specific benefits, and our methods applies to both representations. For the applications we consider in Sec.~\ref{sec:svr}, SDFs appear to represent more accurate surfaces. Occupancy fields are however more suited to union operations in the implicit domain, since the minimum of 2 occupancy fields yields a valid occupancy. This property can be useful for combining shape primitives in Constructive Solid Geometry (CSG) applications~\cite{Fougerolle05}, but does not always hold for SDFs. Similarly, computing ground truth SDF values of a mesh with internal surface elements yields a false zero-levelset with no change of sign, and this source of inaccuracy is removed when using occupancy.}

However, for applications requiring explicit surface parameterizations, the non-differentiability of iso-surface extraction so far largely prevented the exploitation of implicit representations.
Exceptions {are~\cite{Niemeyer20,Yariv20}} that propose a solution to differentiate through iso-surface extraction specifically tailored to differentiable rasterization. By contrast, our method for implicit differentiation is agnostic to the downstream task.
{Our expression is similar to the one of~\cite{Stam11}, which formulates surface derivative with respect to time instead of latent vectors. However, our derivation clarifies the underlying assumptions, namely that the vertices move towards their closest neighbors when the surface deforms infinitesimally. }

\comment{\er{However, for applications requiring explicit surface parameterizations, the non-differentiability of iso-surface extraction has so far prevented additional benefits from training models end-to-end or exploiting the advantages of implicit representations to define effective surface parameterizations.}}

\subsection{Converting Implicit Functions to Surface Meshes}
The Marching Cube (MC) algorithm~\cite{Lorensen87,Newman06} is a popular way to convert implicit functions to surface meshes. The algorithm proceeds by sampling the field on a discrete 3D grid, detecting zero-crossing of the field along grid edges, and building a surface mesh using a lookup table.  Unfortunately, the process of determining the position of vertices on grid edges involves linear interpolation, which does not allow for topology changes through backpropagation~\cite{Liao18a}, as illustrated in Fig.~\ref{fig:mc}(a). Because this is a central motivation for this work, we provide a more detailed analysis of this shortcoming in the supplementary material. In what follows, we discuss two classes of methods that tackle the non-differentiability issue. The first one emulates iso-surface extraction with deep neural networks, while the second one avoids the need for mesh representations by formulating objectives directly in the implicit domain.

\subsubsection{Emulating Iso-Surface Extraction}
In~\cite{Liao18a} Deep Marching Cubes maps voxelized point clouds to a probabilistic topology distribution and vertex locations defined over a discrete 3D Euclidean grid through a 3D CNN. While this allows changes to surface topology through backpropagation, the probabilistic modeling requires keeping track of all possible topologies at the same time, which, in practice, limits resulting surfaces to low resolutions. Voxel2mesh~\cite{Wickramasinghe20} deforms a mesh primitive and adaptively increases its resolution. While this makes it possible to represent high resolution meshes, it prevents changes of topology.

\subsubsection{Writing Objective Functions in terms of Implicit Fields}
In~\cite{Michalkiewicz19}, variational analysis is used to re-formulate standard surface mesh priors, such as those that enforce smoothness, in terms of implicit fields. Although elegant, this technique requires carrying out complex derivations for each new loss function and can only operate on an Euclidean grid of fixed resolution.  The differentiable renderers of~\cite{Jiang20a} and~\cite{Liu20g} rely on sphere tracing and operate directly in terms of implicit fields.
Unfortunately, since it is computationally intractable to densely sample the underlying volume, these approaches either define implicit fields
over a low-resolution Euclidean grid \cite{Jiang20a} or rely on heuristics to accelerate ray-tracing \cite{Liu20g}, while reducing accuracy.
3D volume sampling efficiency can be improved by introducing a  sparse set of anchor points when performing ray-tracing~\cite{Liu19j}. However, this requires reformulating standard surface mesh regularizers in terms of implicit fields using computationally intensive finite differences. Furthermore, these approaches are tailored to differentiable rendering, and are not directly applicable to different settings that require explicit surface modeling, such as computational fluid dynamics. {This also applies to~\cite{Niemeyer20,Yariv20} that use implicit differentiation for implicit surface rendering. Both can be seen as special cases of the gradients we derive  where surface points only move along the viewing direction.}



\section{Method}
\label{sec:method}

Tasks such as Single view 3D Reconstruction (SVR)~\cite{Kanazawa18b,Henderson19} or shape design in the context of CFD~\cite{Baque18} are commonly performed by deforming the shape of a 3D surface mesh $\mathcal{M}=(V,F)$, where $V = \{ \bv_1,\bv_2,... \}$ denotes vertex positions in $\mathbb R^3$ and $F$ facets, to minimize a task-specific loss function $\mathcal{L}_{\text{task}}(\mathcal{M})$. $\mathcal{L}_{\text{task}}$ can be, e.g., an image-based loss defined on the output of a differentiable renderer for SVR or a measure of aerodynamic performance for CFD. 

To perform surface mesh optimization robustly, a common practice is to rely on low-dimensional parameterizations that are either learned~\cite{Blanz99,Park19c,Bagautdinov18} or hand-crafted~\cite{Baque18,Umetani18,Remelli17}. In that setting, a differentiable function maps a low-dimensional set of parameters $\mathbf z$ to vertex coordinates $V$, implying a fixed topology. Allowing changes of topology, an implicit surface representation would pose a compelling alternative but conversely require a \textit{differentiable} conversion to explicit representations in order to backpropagate gradients of $\mathcal{L}_{\text{task}}$.

In the remainder of this section, we first recapitulate neural implicit surface representations that underpin our approach. We then introduce our main contribution, a differentiable approach to computing surface samples and updating their 3D coordinates to optimize $\mathcal{L}_{\textit{task}}$. Finally, we present \dmesh{}, a fully differentiable surface mesh parameterization that can represent arbitrary topologies. 

\subsection{Deep Implicit Field Representation}
\label{sec:rep}

In this work, we represent a generic watertight surface $S$ implicitly by a function $s : \mathbb R ^ 3 \rightarrow \mathbb{R}$. Typical choices for $s$ include the Signed Distance Function (SDF) where $s(\bx)$ is $d(\bx,S)$ if $\bx$ is outside $S$ and $-d(\bx, {S})$ if it is inside, where $d$ is the Euclidean distance; and Occupancy Maps with $s(\bx)=1$ inside and $s(\bx)=0$ outside.

Given a dataset of watertight surfaces $\mathcal D$, such as ShapeNet~\cite{Chang15},  we train a Multi-Layer Perceptron (MLP) $f_{\Theta}$ as in~\cite{Park19a} to approximate $s$ over such set of surfaces $\mathcal D$  by minimizing
\begin{small}
\begin{align}
\mathcal L_{\text{imp}}(\{\mathbf z_S\}_{S \in \mathcal D}, \Theta)  =  \mathcal L_{\text{data}}(\{\mathbf z_S\}_{S \in \mathcal D}, \Theta)  + \lambda_{\text{reg}} \sum _{S \in \mathcal D} \| \bz_S \|_2^2 \; ,  \label{eq:imp}
\end{align}
\end{small}
where $\bz_S \in \mathbb R ^ Z $ is the $Z$-dimensional encoding of surface $S$, $\Theta$ denotes network parameters,  $L_{\text{data}}$ is a data term that measures how similar $f_{\Theta}$ is to the ground-truth function $s$ corresponding to each sample surface, and $\lambda_{\text{reg}}$ is a weight term balancing the contribution of reconstruction and regularization in the overall loss. 

In practice when $s$ is  a signed distance, we take $\mathcal L_{\text{data}}$ to be the $L_1$ loss
\begin{align}
\mathcal L_{\text{data}} = \sum _{S \in \mathcal D}\frac{1}{|X_S|}\sum _{\bx \in X_S}| f_\Theta (\bz_S, \mathbf x) - s(\mathbf x) | \; ,  \label{eq:sdf}
\end{align}
where $X_S$ denotes sample 3D points on the surface $S$ and around it. When $s$ is an occupancy map, we take $\mathcal L_{\text{data}}$ to be the binary cross entropy loss
\begin{align} \label{eq:occ}
	\mathcal L_{\text{data}} = - \sum _{S \in \mathcal D}\frac{1}{|X_S|}\sum _{\bx \in X_S} 
	& s(\bx) \log(f_\Theta (\bz_S, \mathbf x)) \\
	& + (1-s(\bx)) \log( 1- f_\Theta (\bz_S, \mathbf x)) \; . \nonumber
	\end{align}

Once trained, $s$ is approximated by $f_{\Theta}$ which is by construction continuous and differentiable almost everywhere for all standard activation functions (ReLU, sigmoid, tanh...). Consequently, $S$ can be taken to be a level-set of $\{\bx \in \mathbb{R}^3 , f_{\Theta}(\bz_S, \bx) = \alpha\}$, when $\alpha$ is zero for SDFs and typically 0.5 for occupancy grids. Since $f_{\Theta}$ is defined up to a constant, we will refer to zero-crossings in the rest of the paper for simplicity.

\subsection{Differentiable Iso-Surface Extraction}
\label{sec:diff}


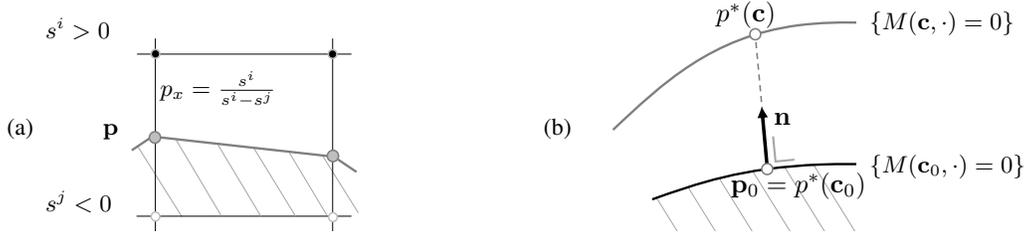
\begin{figure*}[t]
\vspace{-6pt}
            \begin{center}
			\begin{overpic}[clip, trim=2.0cm 10cm 10cm 4cm,width=0.7\textwidth]{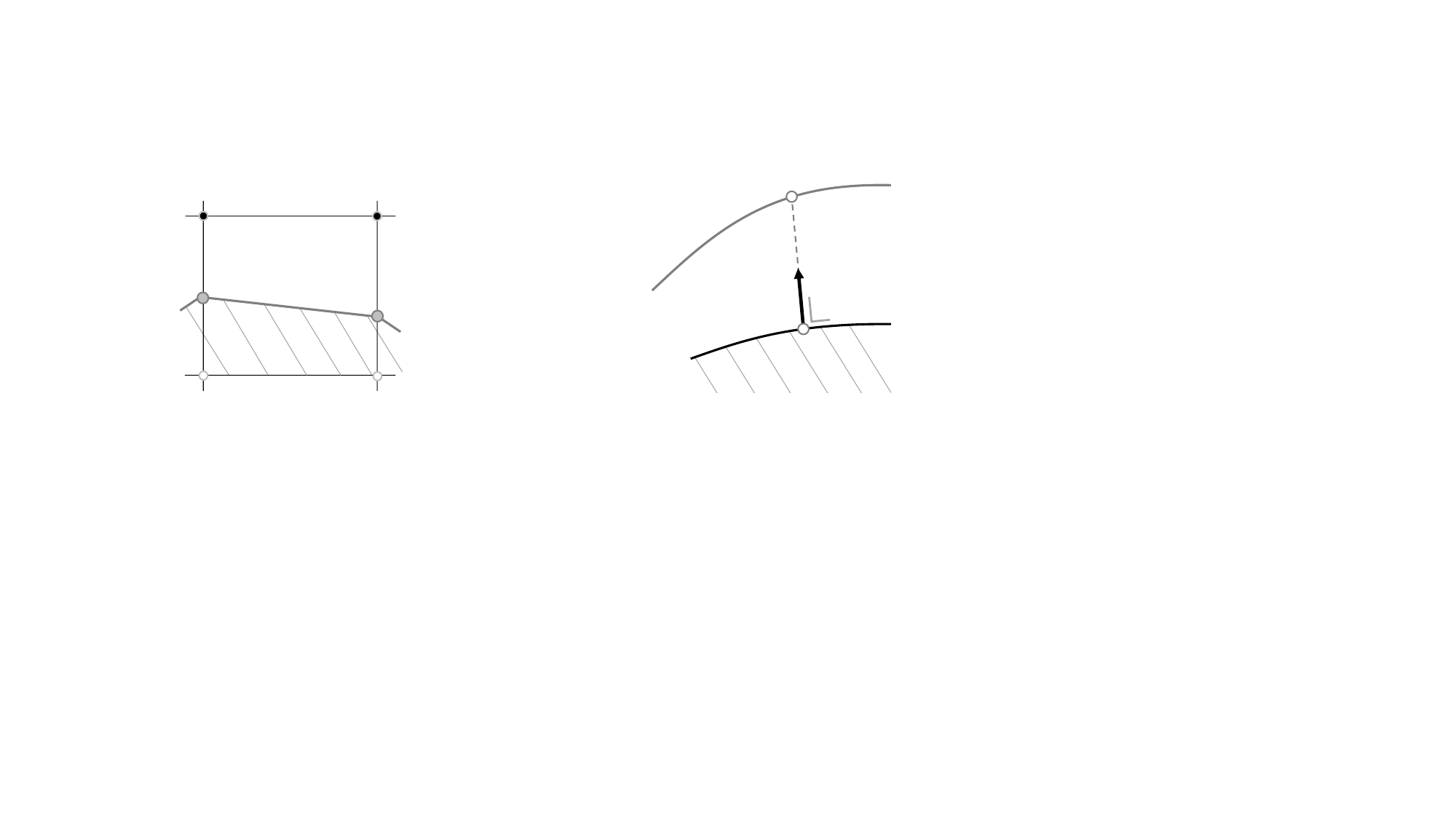}
			\put(1,20){\small{$ s ^i > 0 $}}
			\put(1, 2){\small{$ s ^j < 0 $}}
			\put(7,10){\small{$ \bp $}}
			\put(13,14){\small{$ p_x = \frac{s ^i}{s ^i - s ^j } $}}
			
			\put(87,6){\small{$\{M(\bc_0, \cdot ) = 0\} $}}
			\put(87,21){\small{$\{M(\bc, \cdot ) = 0\} $}}
			\put(72.5,4){$\bp_0 = p^*(\bc_0)$}
			\put(77,11){$\bn$}
			\put(71,21.8){$p^*(\bc)$}
			
			\put(-3,10){\small{(a)}}
			\put(53,10){\small{(b)}}

			\end{overpic} 
			\end{center}
	\vspace{-3mm}
	\caption{\textbf{Marching Cubes differentiation vs Iso-surface differentiation.} (a) Marching Cubes determines the position $p_x$ of a vertex $\bp$ along an edge via linear interpolation. This does not allow for effective back-propagation when topology changes because its behavior is degenerate when $s ^i=s ^j$ as shown in~\cite{Liao18a}. (b) Instead, we adopt a \textit{continuous} model expressed in terms of how {implicit} function perturbations locally impact surface geometry. Here, we depict the geometric relation between {implicit parameter perturbation $\bc_0 \hookrightarrow \bc$ and local surface change $\bp_0 \hookrightarrow p^*(\bc)$, which we exploit to compute $\frac{\partial p^*(\bc) }{\partial \bc}$ } even when the topology changes.
	}
		\label{fig:mc}
\end{figure*}

Once the weights $\Theta$ of Eq.~\ref{eq:imp} have been learned, $f_{\Theta}$ maps a latent vector $\bz$ to a signed distance or occupancy field and the surface of interest is its zero level set. Recall that our goal is to minimize the objective function $\mathcal{L}_{\text{task}}$ introduced at the beginning of this section.  As it takes as input a mesh defined in terms of its vertices and facets, evaluating it and its derivatives requires a \textit{differentiable} conversion from an implicit field to a set of vertices and facets, something that Marching Cubes does not provide, as depicted by Fig.~\ref{fig:mc}(a). More formally, we need to evaluate
\begin{align}
\frac{\partial \mathcal L_{\text{task}}}{\partial \bc} &= \sum_{\bx \in V}\frac{\partial \mathcal{L}_{\text{task}}}{\partial \bx}  \frac{\partial \bx}{\partial \bc} \; ,
\label{eq:chainRule}
\end{align}
where the $\bx$ are mesh vertices and therefore on the surface. $\bc$ stands for either the latent $\bz$ vector if we wish to optimize $\mathcal L_{\text{task}}$ with respect to $\bz$ only or for the concatenation of the latent vector and the network weights $[\bz | \Theta]$ if we wish to optimize with respect to both the latent vectors and the network weights. Note that we compute ${\partial \mathcal L_{\text{task}}} / {\partial \bc}$ by summing over the mesh vertices but we could use any other sampling of the surface.  

\subsubsection{Differentiating the Loss Function}

In this work, we take inspiration from classical functional analysis~\cite{Allaire02} and reason about the {\it continuous} zero-crossing of the implicit function $s$ rather than focusing on how vertex coordinates depend on the implicit field $f_\Theta$ when sampled by the marching cubes algorithm. To this end, we prove below that
\begin{theorem}
	\label{th:shapeDerivative}
	If the gradient of $f_{\Theta}$ at point $\bx$ located on the surface does not vanish, then $\frac{\partial \bx}{\partial \bc} = - \tfrac{\bn }{\left \| \bn  \right \|^2} \frac{\partial f_{\Theta}(\bz, \bx)}{\partial \bc}$ where $\bn = \nabla f_{\Theta}(\bx)$ is the normal to the surface at $\bx$.
\end{theorem}
Injecting this expression of ${\partial \bx} / {\partial \bc}$ into Eq.~\ref{eq:chainRule} yields
\begin{align}
\label{eq:backward}
\frac{\partial \mathcal{L}_{\text{task}}}{\partial \bc} = - \sum_{\bx \in V} \frac{\partial \mathcal{L}_{\text{task}}}{\partial \bx}  \frac{\nabla f_{\Theta} }{\left \| \nabla f_{\Theta}  \right \|^2} \frac{\partial f_{\Theta}}{\partial \bc} \; .
\end{align}
Note that when $s$ is an SDF, $\| \nabla s \| = 1$ and therefore $\| \nabla  f_{\Theta} \| \approx 1$. The $\| \nabla  f_{\Theta} \| $ term from Eq.~\ref{eq:backward}  can then be ignored, which is consistent with the result we presented in~\cite{Remelli20b}.


\vspace{2mm}\noindent {\bf Proof of Theorem 1.}
It starts with the well known
\begin{theorem}[\bf Implicit Function Theorem - IFT]
  Let $F: \mathbb{R}^m  \times \mathbb{R}^n \rightarrow  \mathbb{R}^n$ and $\bc_0 \in \mathbb{R}^m , \bp_0 \in \mathbb{R}^n$ such that:
  \begin{enumerate}
    \item $F(\bc_0, \bp_0) = 0$ ;
    \item $F$ is continuously differentiable in a neighborhood of $(\bc_0, \bp_0)$ ;
    \item the partial Jacobian $\partial_p F(\bc_0, \bp_0) \in \mathbb{R}^{n \times n}$ is non-singular.
  \end{enumerate}
  Then there exists a \textbf{unique} differentiable function $p^*: \mathbb{R}^m \rightarrow \mathbb{R}^n$ such that:
  \begin{enumerate}
    \item $\bp_0 = p^*(\bc_0)$ ;
    \item $F(\bc, p^*(\bc))=0$ for all $\bc$ in the above mentioned neighborhood of $\bc_0$ ;
    \item $\partial p^*(\bc_0) = - \left [ \partial_p F(\bc_0, \bp_0) \right ]^{-1} \partial_c F(\bc_0, \bp_0)$, that is,  a matrix in $\mathbb{R}^{n \times m}$.
  \end{enumerate}
\end{theorem}
Intuitively, $p^*$ returns the solutions of a system of $n$ equations---the $n$ output values of $F$---with $n$ unknowns. For our purposes, $\bc \in \mathbb{R}^m$ can be either the shape code and the network weights jointly or the shape code only, as discussed above.

To apply the IFT to our problem, let us rewrite $f_{\Theta}$ as a function $M: \mathbb{R}^m  \times \mathbb{R}^3 \rightarrow  \mathbb{R}$ that maps $\bc \in \mathbb{R}^m$ and a point in $\bp \in \mathbb{R}^3$ to a scalar value $M(\bc, \bp) \in \mathbb{R}$. The IFT does not directly apply to $M$ because it operates from $\mathbb{R}^m \times \mathbb{R}^3$ into $\mathbb{R}$ instead of into $\mathbb{R}^3$. Hence, we must add two more dimensions to the output space of $M$.

To this end, let $\bc_0 \in \mathbb{R}^m$; $\bp_0 \in \mathbb{R}^3$ such that $M(\bc_0, \bp_0) = 0$, {meaning that $\bp_0$ is on the implicit surface defined by parameter $\bc_0$; and $\mathbf{u} \in \mathbb{R}^3$ and $\mathbf{v} \in \mathbb{R}^3$ such that $(\mathbf{u}, \mathbf{v})$ is a basis of the tangent plane to the surface $\{M(\bc_0, \cdot ) = 0\}$ at $\bp_0$.} Let $\bn= \partial_p M(\bc_0,\bp_0)$ be the normal vector to the surface at $\bp_0$. This lets us define the function $F: \mathbb{R}^m  \times \mathbb{R}^3 \rightarrow  \mathbb{R}^3$ as 
\begin{align}
  F(\bc,\bp) \mapsto  \biggl(\begin{smallmatrix} M(\bc, \bp)\\ 
  (\bp - \bp_0) \cdot \mathbf{u} \\ 
  (\bp - \bp_0) \cdot \mathbf{v}
  \end{smallmatrix}\biggr) \; ,
\end{align} 
By construction, we have $\mathbf{n} \cdot \mathbf{u} = \mathbf{n} \cdot \mathbf{v} = 0$ and $F(\bc_0, \bp_0)=0$.

Note that the first value of the $F(\bc,\bp)$ vector is zero when the point $\bp$ is on the surface defined by $\bc$ while the other two are equal to zero when $(\bp - \bp_0)$ is perpendicular to the surface defined by $\bc_0$. By zeroing all three, $p^*$ returns a point $\bp$ that is on the surface for $\bc \neq \bc_0$ and such that $(\bp - \bp_0)$ is perpendicular to the surface. A geometric interpretation is that {$\bp_0$ is the point on the surface defined by $\bc_0$ that is the closest to $p^*(\bc)$. This is illustrated on Fig.~\ref{fig:mc}(b).}

Given the IFT applied to F, there is a mapping $p^*: \mathbb{R}^m \rightarrow \mathbb{R}^3$ such that
\begin{enumerate}
  \item $\bp_0 = p^*(\bc_0)$ ;
  \item $F(\bc, p^*(\bc))= \biggl(\begin{smallmatrix}
    0\\ 
    0 \\ 
    0
    \end{smallmatrix}\biggr)$ for all $\bc$ in a neighborhood of $\bc_0$ .

\item  $\partial p^*(\bc_0) = - \left [ \partial_p F(\bc_0, \bp_0) \right ]^{-1} \partial_c F(\bc_0, \bp_0)$. 
\end{enumerate}
We have
\begin{align}
  \partial_c F(\bc_0, \bp_0) &= \biggl(\begin{smallmatrix}
    \partial_c M(\bc_0, \bp_0)\\ 
    0\\ 
    0
    \end{smallmatrix}\biggr) \in \mathbb{R}^{3\times m} \; , \\
 \partial_p F(\bc_0, \bp_0)   &= \biggl(\begin{smallmatrix}
     \bn \\ \bu \\ \bv
    \end{smallmatrix}\biggr) \in \mathbb{R}^{3\times 3}
\end{align}
Given that the last two rows of $\partial_c F(\bc_0, \bp_0)$ are zero, to compute $\partial p^*(\bc_0)$ according to the IFT, we only need to evaluate the first column of  $\left [ \partial_p F(\bc_0, \bp_0) \right ]^{-1}$. As the two last rows of $ \partial_p F(\bc_0, \bp_0)$ are $\bu$ and $\bv$ that are unit vectors such that $\bu \cdot \bv = \mathbf{n} \cdot \mathbf{u} = \mathbf{n} \cdot  \mathbf{v} = 0$, that first column has to be ${\bn} / { \| \mathbf{n} \|^2}$.
Hence, we have
\begin{align}
  \partial p^*(\bc_0) 
  & = - \left [ \partial_p F(\bc_0, \bp_0) \right ]^{-1} \partial_c F(\bc_0, \bp_0) \; , \\
  & = - 
  \tfrac{\mathbf{n} }{\left \| \mathbf{n}  \right \|^2} \times \partial_c F(\bc_0, \bp_0) 
   \in \mathbb{R}^{3\times m} \; . \nonumber
\end{align}
Recall that $p^*$ maps a code $\bc$ {in the neighborhood of $\bc_0$ to a 3D point such that $M(\bc,p^*(\bc)) = f_{\Theta}(p^*(\bc), \bz)= 0$. In other words, $\bp_0=p^*(\bc_0)$ }is a point on the implicit surface defined by $f_{\Theta}$ when $\bc = \bc_0$ and we have
\begin{align}
\frac{\partial \bp_0}{\partial \bc} 
& = - \tfrac{\mathbf{n} }{\left \| \mathbf{n}  \right \|^2} \times \partial_c F(\bc_0, \bp_0)  \; , \\
& = - \tfrac{\mathbf{n} }{\left \| \mathbf{n}  \right \|^2} \times  \frac{\partial f_{\Theta}(\bz, \bp_0)}{\partial \bc}
\end{align}
{where $\bc$ either stands for the latent vector $\bz$ or the concatenation of the latent vector and the network weights $[\bz|\Theta]$.} \hfill \ensuremath{\Box} 

\subsubsection{Forward and Backward Passes}
\label{subsec:forward_backward}
Recall that the goal of our forward pass is to extract surface mesh $\mathcal M = (V,F)$ from an underlying neural implicit field $f_\Theta$. Because sampling a DIF on a dense Euclidean Grid is computationally intensive, we use a hierarchical approach to reduce the total number of evaluations during the forward and backward passes summarized by Algorithms~\ref{algo:mesh-fwd} and~\ref{algo:mesh-back}. 

We start by evaluating $f_\Theta$ on a low resolution grid, and then iteratively subdivide each voxel and re-evaluate the DIF only where needed until we reach a desired grid resolution, as in~\cite{Mescheder19, Venkatesh21}. When our DIF is a signed distance function, we subdivide voxels if the field absolute value on any of the voxel corners $\{| f_\Theta ( \bx _i) | \}_{i=1} ^8$ is smaller than the voxel diagonal $\sqrt{2} \Delta x$, where $\Delta x$ denotes voxel size. When it is an occupancy map, we only  split voxels when the occupancy map does not have the same value at all corners. For this to work well,  we have to start from a grid that roughly captures the object topology to make hierarchical iso-surface extraction converge. In practice, we have found that starting with a {$32^3$} grid is enough.

In this way, we can quickly obtain a high resolution DIF grid without needless computation far away from the surface.  Once the grid has been assembled, we use a GPU-accelerated marching cubes algorithm~\cite{Yatagawa20} to extract the vertices and vertex normals needed to perform the backward pass. The backward pass then performs the computation of Eq.~\ref{eq:backward}. This requires computing the values of $f_\Theta(\bz,\bv)$ and its derivatives $\frac{ \partial f_{\Theta} }{ \partial \bc}(\bz,\bv)$ at the newly found vertices $\bv$. We show that the resulting overhead is small in the results section.


\begin{center}
\begin{minipage}[t]{0.38\textwidth}
\begin{algorithm}[H]

    \caption{\dmesh{} Forward}\label{algo:mesh-fwd}
    \begin{algorithmic}[1]
        \State \text{\textbf{input:} latent code $\bz$, DIF weights $\Theta$ } 
        \State \text{\textbf{output:} surface mesh $\mathcal M = (V,F)$} 
        
        \State \text{assemble coarse 3D grid $G$} 
        \State \text{sample field on grid $S = f_\Theta(\bz, G)$} 
        \State while $G$ has not reached target resolution:
        \State \hspace{14pt}$G_s$ = split($G$)
        \State  \hspace{14pt}$S$ = $S$ + $f_\Theta(\bz, G_s)$
        \State \hspace{14pt}$G$ = $G$ + $G_s$ 
        \State \text{extract iso-surface $(V,F) = $ MC$(S)$} 
        \State \textbf{Return}  $\mathcal M = (V,F)$
    \end{algorithmic}
\end{algorithm}
\begin{algorithm}[H]
    \caption{\dmesh{} Backward}\label{algo:mesh-back}
    \begin{algorithmic}[1]
        \State \text{\textbf{input:} upstream gradient $\frac{\partial \mathcal L}{\partial \bv }$ for $\bv \in V$} 
        \State \text{\textbf{output:} downstream gradient $\frac{\partial \mathcal L}{\partial \bc}$} 
        \State \text{$\frac{\partial \mathcal L}{\partial f_\Theta}(\bv) = - \frac{\partial  \mathcal L}{\partial \bv} \; \tfrac{\mathbf{n} }{\left \| \mathbf{n}  \right \|^2} $ for $\bv \in V$} 
        \State extra pass on samples $\frac{ \partial f_{\Theta} }{ \partial \bc}(\bz,\bv)$
        \State \textbf{Return}  $\frac{\partial \mathcal L}{\partial \bc} = \sum_{\mathbf v \in V} \frac{\partial \mathcal L}{\partial f_\Theta}(\bv) \frac{\partial f_\Theta}{\partial \bc}(\bv)$
    \end{algorithmic}
\end{algorithm}
\end{minipage}
\end{center}




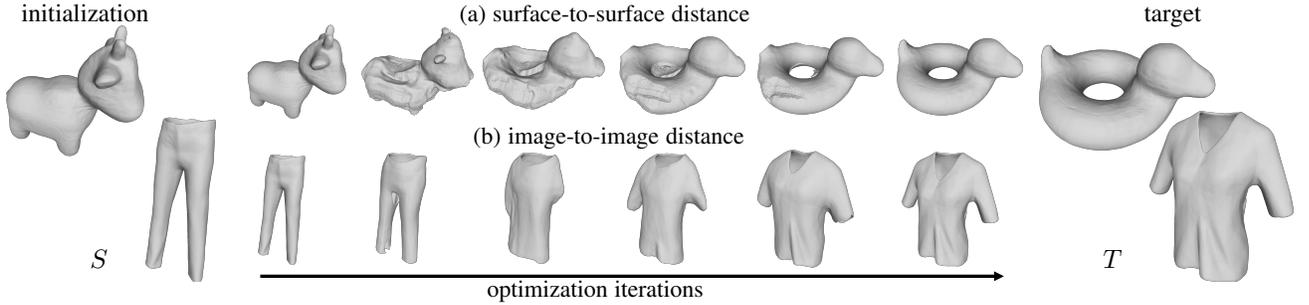
\begin{figure*}[t]
		\hspace{-2pt}
		\begin{center}
			\begin{overpic}[clip, trim=2.2cm 7.0cm 2.2cm 6.5cm,width=1.0\textwidth]{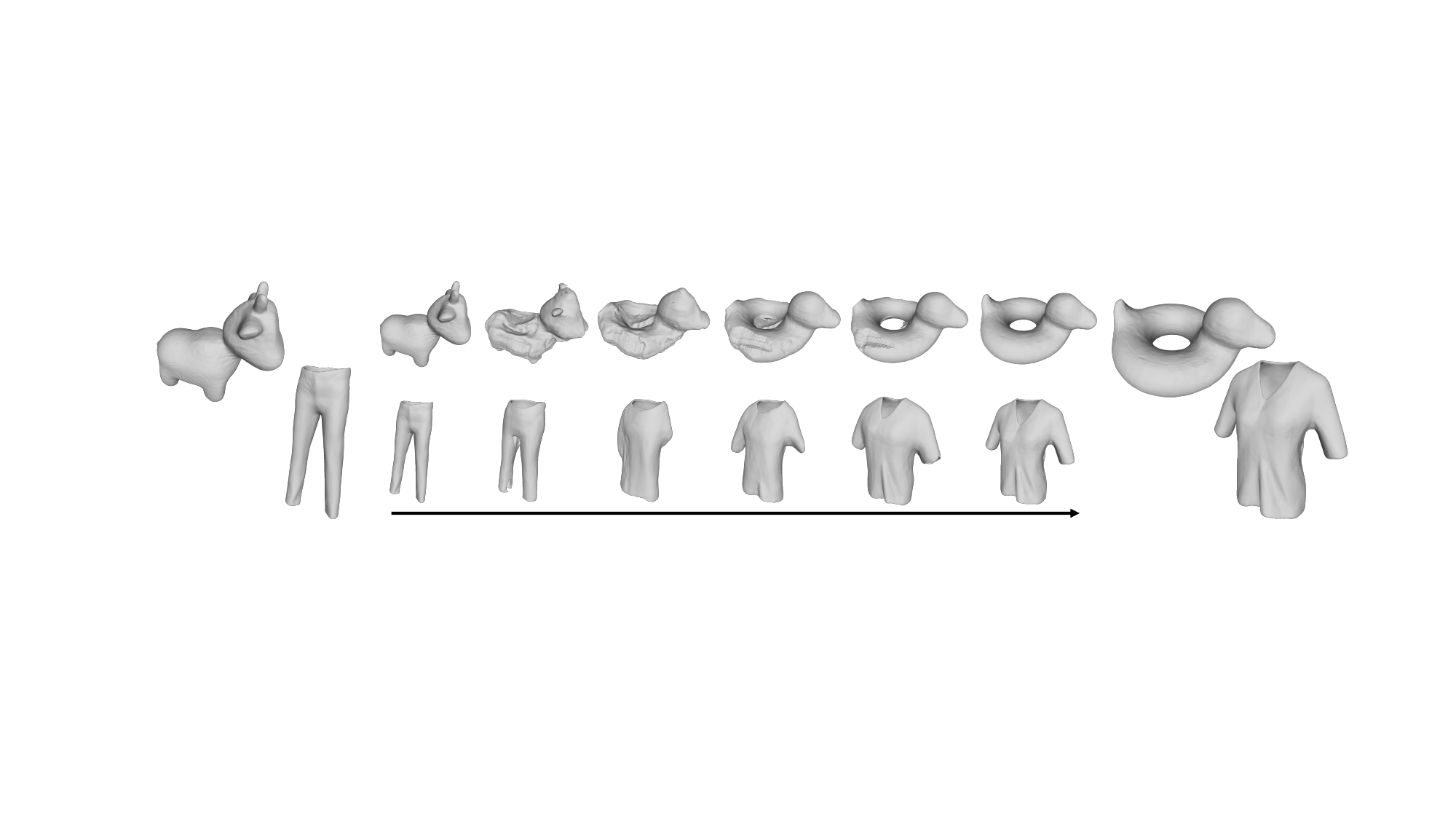}
			\put(40,-1.1){\small{optimization iterations}}
			
			\put(38,19){\small{(a) surface-to-surface distance }}
			\put(39,10.1){\small{(b) image-to-image distance }}
			
			\put(6,19){{initialization }}
			\put(88,19){{target}}
			
			\put(11,1){{$S$}}
			\put(85,1){{$T$}}
			
			\end{overpic}
		\end{center}
		\vspace{-10pt}
		\caption{\textbf{Topology-Variant Parameterization.}
 We minimize (a) a surface-to-surface or (b) an image-to-image distance with respect to the latent vector $\bz$ to transform {an initial shape into a target one that has a different genus}.
 This demonstrates that we can backpropagate gradient information from mesh vertices to latent vector while modifying surface mesh topology.}
\label{fig:sanity}
\vspace{-6pt}
\end{figure*}

\section{Experiments}
\label{sec:exp}

We first use synthetic examples to show that, unlike marching cubes, our approach allows for differentiable topology changes. We then demonstrate that we can exploit surface mesh differentiability to outperform state-of-the-art approaches on three very different tasks, Single view 3D Reconstruction, Aerodynamic Shape Optimization, and Full Scene 3D Reconstruction from Scans. In these experiments, we use Theorem~\ref{th:shapeDerivative} with $\bc=\bz$, that is, we only optimize with respect to shape codes while keeping the network weights fixed.
In the final subsection, we discuss an application in which we take {$\bc=\Theta$}, that is, we optimize with respect to the network weights.

\subsection{Differentiable Topology Changes}

In the experiment depicted by Fig.~\ref{fig:sanity} we train two separate networks $f_{\Theta_1}$ and $f_{\Theta_2}$ that implement the approximate implicit field of Eq.~\ref{eq:imp}. $f_{\Theta_1}$ is a deep occupancy network trained to minimize the loss of Eq.~\ref{eq:sdf} on two models of a cow and a rubber duck. They are of genus 0 and 1, respectively.  $f_{\Theta_2}$ is a deep signed distance function network trained to minimize the loss of Eq.~\ref{eq:occ} on four different articles of clothing, a t-shirt, a pair of pants, a dress, and a sweater. Note that the clothes are represented as open surface meshes without inside/outside regions. Hence, they are not watertight surfaces. To nevertheless represent them using a signed distance function, we first compute unsigned distances to the surfaces, subtract a small $\epsilon=0.01$ value and treat the result as a signed distance function. This amounts to representing the garments as watertight thin surfaces of thickness $2\epsilon$. 
{We visualize additional watertight reconstructions in the supplementary material.}

In short, $f_{\Theta_1}$ associates to a latent vector $\bz$ an implicit field $f_{\Theta_1}(\bz)$ that represent a cow, a duck, or a mix of the two, while $f_{\Theta_2}$ associates to $\bz$ a garment representation that can be a mixture of the four it was trained on. In Fig.~\ref{fig:sanity}, we start from a shape $S$ and find a vector $\bz$ so that $f_{\Theta_x}(\bz)$ with $x \in{1,2}$ approximates $S$ as well as possible. We then use the pipeline of Sec.~\ref{sec:diff} to minimize a differentiable objective function of $\bz$ so that $f_{\Theta_x}(\bz)$ becomes an approximation of a different surface $T$. 

When using $f_{\Theta_1}$, we take the differentiable objective function to be minimized to be the chamfer distance between the current surface $C$ and the target surface $T$
\begin{equation}
\mathcal{L}_{\text{task1}}(C,T)  = \min _{c\in C} d(c,T) + \min _{t \in T} d(C,t) \; .
\end{equation}
where $d$ is the point-to-surface distance in 3D. When  using $f_{\Theta_2}$, we take it to be
\begin{equation}
    \mathcal{L}_{\text{task2}}(C,T)   =\| \text{DR}(C)- \text{DR}(T)\|_1 \; , 
\end{equation}
where $\text{DR}$ is the output of a differentiable rasterizer~\cite{Kato18}.  In other words, $\mathcal{L}_{\text{task1}}$ is the surface-to-surface distance while $\mathcal{L}_{\text{task2}}$ is the image-to-image $L_1$ distance between the two rendered surfaces.

In both cases, the left shape smoothly turns into the right one, and changes its genus to do so. Note that even though we rely on a deep implicit field to represent our topology-changing surfaces, unlike in~\cite{Michalkiewicz19,Jiang20a,Liu20g,Liu19j},  we did {\it not} have to reformulate the loss functions in terms of implicit surfaces.

\subsection{Single view 3D Reconstruction}
\label{sec:svr}

Single view 3D Reconstruction (SVR) has emerged as a standardized benchmark to evaluate 3D shape representations~\cite{Choy16b,Fan17a,Groueix18a,Wang18e,Chen19c,Mescheder19,Pontes18,Gkioxari19,Richter18,Xu19b,Tatarchenko19}. We demonstrate that it is straightforward to apply our approach to this task on two standard datasets, ShapeNet~\cite{Chang15} and Pix3D~\cite{Sun18b}. 

\subsubsection{Differentiable Meshes for SVR.}

As in~\cite{Mescheder19,Chen19c}, we condition our deep implicit field architecture on the input images via a residual image encoder~\cite{He16a}, which maps input images to latent code vectors $\bz$. These latent codes are then used to condition the architecture of Sec.~\ref{sec:rep} and compute the value of deep implicit function $f_{\Theta}$. Finally, we minimize $\mathcal L_{\text{imp}}$ (Eq.~\ref{eq:imp}) wrt. $\Theta$ on a training set of image-surface pairs {generated on the ShapeNet Core~\cite{Chang15} dataset for the cars and chairs object classes. Each object class is split into 1210 training and 112 testing shapes, each of which is paired with the renderings provided in~\cite{Xu19b}. 3D supervision points are generated according to the procedure of~\cite{Park19c}.} {To showcase that our differentiability results work with \textit{any} implicit representation, we train networks that output either signed distance fields or occupancy fields. To this end, we minimize the loss functions Eqs.~\ref{eq:sdf} and~\ref{eq:occ}, respectively. }

We begin by using the differentiable nature of our mesh representation to refine the output of an encoder, as depicted by the top row of Fig.~\ref{fig:teaser}. As in many standard methods, we use our encoder to predict an initial latent code $\bz$. Then, unlike in standard methods, we refine the predicted shape $\mathcal M$, that is, given the camera pose associated to the image and the current value of $\bz$, {we project the reconstructed mesh back to the image plane so that the projection matches the object silhouette $\cal S$ in the image as well as possible. } To this end, we define the task-specific loss function $L_{task}$ to be minimized, as discussed in Section~\ref{sec:method}, in one of two ways:
\begin{align}
\mathcal{L}_{\text{task3}} &= \| \text{DR}_{\text{silhouette}}(\mathcal M(\bz))- T \|_1 \:,      \label{eq:rasterLoss} \\
\mathcal{L}_{\text{task4}} &= \sum_{a \in A} \min_{b \in B} \left \| a - b \right \|^2 + \sum_{b \in B} \min_{a \in A} \left \| a - b \right \|^2 \: .\label{eq:chamfLoss} 
\end{align}
In Eq.~\ref{eq:rasterLoss}, $T$ denotes the silhouette of the target surface and $\text{DR}_{\text{silhouette}}$ is the differentiable rasterizer of~\cite{Kato18} that produces a binary mask from the mesh generated by the latent vector $\bz$.   In Eq.~\ref{eq:chamfLoss}, $A\subset [-1,1]^2$ denotes the 2D coordinates of $T$'s external contour while $B \subset [-1,1]^2$ denotes those of the external contour of $\mathcal M(\bz)$. We refer the interested reader to~\cite{Guillard21} for more details on this objective function. Note that, unlike that of $\mathcal{L}_{\text{task3}}$, the computation of $\mathcal{L}_{\text{task4}}$ does not require a differentiable rasterizer. 


\begin{table}[t]
	\caption{\textbf{SVR ablation study on ShapeNet Core.} We exploit end-to-end differentiability to perform image-based refinement using either occupancy maps (Model=Occ.) or signed distance functions (Model=SDF). We report 3D Chamfer distance (Metric=CHD) and normal consistency (Metric=NC) for raw reconstructions (Refine=None),  refinement via differentiable rendering (Refine=DR) and contour matching (Refine=CHD). }
	\vspace{-4mm}
	\label{tab:svr_ablation}
	\begin{small}
	\begin{center}
			\begin{tabular}{c|cc|cc}
				\Xhline{2\arrayrulewidth}
				Metric & Model & Refine & car & chair \\
				\Xhline{2\arrayrulewidth}
				\multirow{6}{*}{\footnotesize \hspace{-2mm} CHD $\!\cdot 10^4$ $\downarrow$} \hspace{-2mm} && None & 3.02 & 11.18\\
				&Occ. & DR & 2.86 ($\downarrow 5.3 \%$) & 10.92 ($\downarrow 2.3 \%$)\\
				&& CHD & 2.65 ($\downarrow 12.3 \%$) & 10.35 ($\downarrow 7.4 \%$)\\
				\cline{2-5}
				&& None & 2.96 &9.07\\
				&SDF& DR &2.73 ($\downarrow 7.8 \%$)&8.83 ($\downarrow 2.6 \%$) \\
				&& CHD & \textbf{2.56} ($\downarrow 13.5 \%$) &\textbf{8.22} ($\downarrow 9.4 \%$) \\
				\Xhline{1\arrayrulewidth}				
				\multirow{6}{*}{\footnotesize \hspace{-2mm} NC $\%$ $\uparrow$ \hspace{-2mm} }&& None & 92.17 & 77.26\\
				&Occ.& DR & 92.07 ($\downarrow 0.1 \%$)& 78.98 ($\uparrow 2.2 \%$)\\
				&& CHD & 92.36 ($\uparrow 0.2 \%$)& 78.49 ($\uparrow 1.6 \%$)\\
				\cline{2-5}
				&& None & 92.29 &78.74\\
				&SDF& DR &92.22 ($\downarrow 0.1 \%$)&80.02 ($\uparrow 1.6 \%$) \\
				&& CHD & \textbf{92.56} ($\uparrow 0.3 \%$) &\textbf{80.17} ($\uparrow 1.8 \%$) \\
				\Xhline{2\arrayrulewidth}				
		\end{tabular}
	\end{center}
	\end{small}
\end{table}

{Recall that we can use either signed distance functions or occupancy fields to model objects. To compare these two approaches, we ran 400 gradient descent iterations using Adam~\cite{Kingma14a} to minimize either $\mathcal{L}_{\text{task3}}$ or $\mathcal{L}_{\text{task4}}$ with respect to $\bz$. This yields four possible combinations of model and loss function and we report their respective performance in Tab.~\ref{tab:svr_ablation}}. They are expressed in terms of two metrics:
\begin{itemize}
    \item The 3D Chamfer distance for 10000 points on the reconstructed and ground truth surfaces, in the original ShapeNet Core scaling. The lower, the better.
    \item A normal consistency score in image space computed by averaging cosine similarities between reconstructed and ground truth rendered normal maps from 8 regularly spaced viewpoints. The higher, the better.
\end{itemize}
All four configurations deliver an improvement in terms of both metrics. However, the combination of using a signed distance field and minimizing the {2D} chamfer distance of $\mathcal{L}_{\text{task4}}$ delivers the largest one. We will therefore refer to it as {\it DeepMesh} and use it in the remainder of this section, unless otherwise specified.


\begin{figure}[t]
\centering
\includegraphics[width=0.49\textwidth, clip, trim=0 0.9cm 0 0]{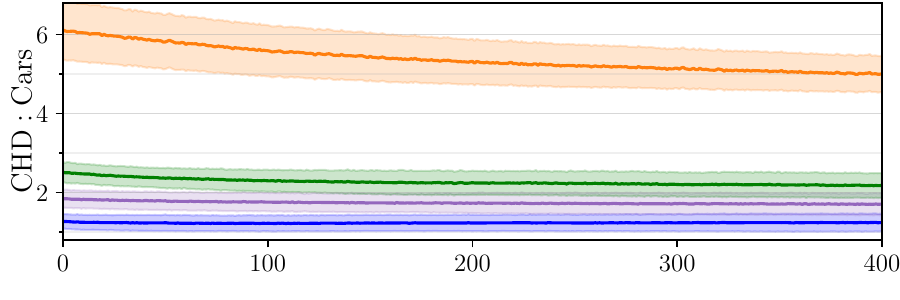}
\includegraphics[width=0.49\textwidth, clip, trim=0 0 0 -0.3cm]{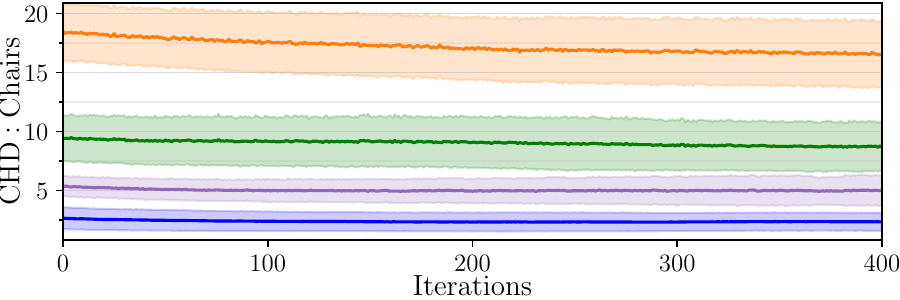}
\vspace{-20pt}
\caption{{\textbf{CHD improvement} over the 400 refinement iterations of \textit{DeepMesh} {for cars (top) and chairs (bottom)}, grouped by quartile in the initial CHD value (from orange = worse 25\% of initial shapes, to blue = top 25\%)}. {Here DeepMesh uses an SDF and refines the contour matching.}}
\label{fig:chd_improvement}
\vspace{-3pt}
\end{figure}

{In Fig.~\ref{fig:chd_improvement} we show the Chamfer distance changing over the 400 refinement iterations of {\it DeepMesh} on {both car and chair categories}. We group the test shapes into quartiles according to their initial Chamfer distance with their corresponding ground truth mesh, and compute the average of each quartile. The Chamfer distance is mostly improved for shapes that have a high initial reconstruction error. For the 3 quartiles that have the best initial reconstruction accuracy, the CHD decrease is smaller and mostly takes place during the first iterations. Although the decrease is small for the first quartile, there still is an improvement from $1.27$ to $1.24$ {for cars, and from $2.65$ to $2.35$ for chairs.}
}

\subsubsection{Comparative Results on ShapeNet}

In Tab.~\ref{tab:shapenet}, we compare our approach against state-of-the-art reconstruction approaches of watertight meshes: Generating surface meshes with fixed topology~\cite{Wang18e}, generating meshes from voxelized intermediate representations~\cite{Gkioxari19}, and representing surface meshes using signed distance functions~\cite{Xu19b}. We used the standard train/test splits and renderings described above for all benchmarked methods.


\begin{table}[ht]
    \caption{\textbf{SVR comparative results on ShapeNet Core.} Exploiting end-to-end differentiability to perform image-based refinement allows us to outperform state-of-the-art methods in terms of both 3D Chamfer distance (CHD) and normal consistency (NC).}
    \vspace{-4mm}
	\label{tab:shapenet}
	\begin{center}
		
			\begin{tabular}{c|c|cc}
				\Xhline{2\arrayrulewidth}
				Metric & Method& car & chair \\
				\Xhline{2\arrayrulewidth}
				\multirow{5}{*}{CHD $\cdot 10^4$ $\downarrow$}&Mesh R-CNN~\cite{Gkioxari19}&4.55&11.13 \\
                &Pixel2Mesh~\cite{Wang18e}&4.72&12.19 \\
                &DISN~\cite{Xu19b}&3.59&8.77 \\
				\cline{2-4}
				&DeepMesh (raw) &2.96 &9.07\\
				&DeepMesh &\textbf{2.56}  &\textbf{8.22} \\
				\Xhline{1\arrayrulewidth}
				\multirow{5}{*}{NC $\%$ $\uparrow$}&Mesh R-CNN~\cite{Gkioxari19}& 89.09 & 74.82\\
				&Pixel2Mesh~\cite{Wang18e}& 89.00 & 72.21 \\
				&DISN~\cite{Xu19b}&91.73&78.58 \\
				\cline{2-4}
				&DeepMesh (raw) &92.29 &78.74\\
				&DeepMesh &\textbf{92.56} &\textbf{80.17} \\
				\Xhline{2\arrayrulewidth}
		\end{tabular}
	\end{center}
\end{table}

{\it DeepMesh} (raw) refers to reconstructions obtained using our encoder-decoder architecture based on signed distance fields but without refinement, which is similar to those of~\cite{Mescheder19,Chen19c}, without any further refinement,  whereas {\it DeepMesh} incorporates the final refinement that the differentiability of our approach allows. {\it DeepMesh} (raw) performs comparably to the other methods whereas  {\it DeepMesh} does consistently better. In other words, the improvement can be ascribed to the refinement stage as opposed to differences in network architecture.  {We provide additional results and describe failure modes in the supplementary material.} 

\subsubsection{Comparative results on Pix3D.}

Whereas ShapeNet contains only rendered images, Pix3D~\cite{Sun18b} is a test dataset that comprises \comment{10069}real images paired to \comment{395 unique} 3D models. Here, we focus on the chair object category and discard truncated images to create a test set of 2530 images. We use it to compare our method with our best competitor~\cite{Xu19b} according to Tab.~\ref{tab:shapenet}. To this end, we use the same networks as for ShapeNet, that is, we do not fine-tune the models on Pix3D images. Instead, we train them only on synthetic chair renderings so as to encourage the learning of stronger shape priors. Testing these networks on real images introduces a large domain gap because synthetic renderings do not account for complex lighting effects or variations in camera intrinsic parameters.

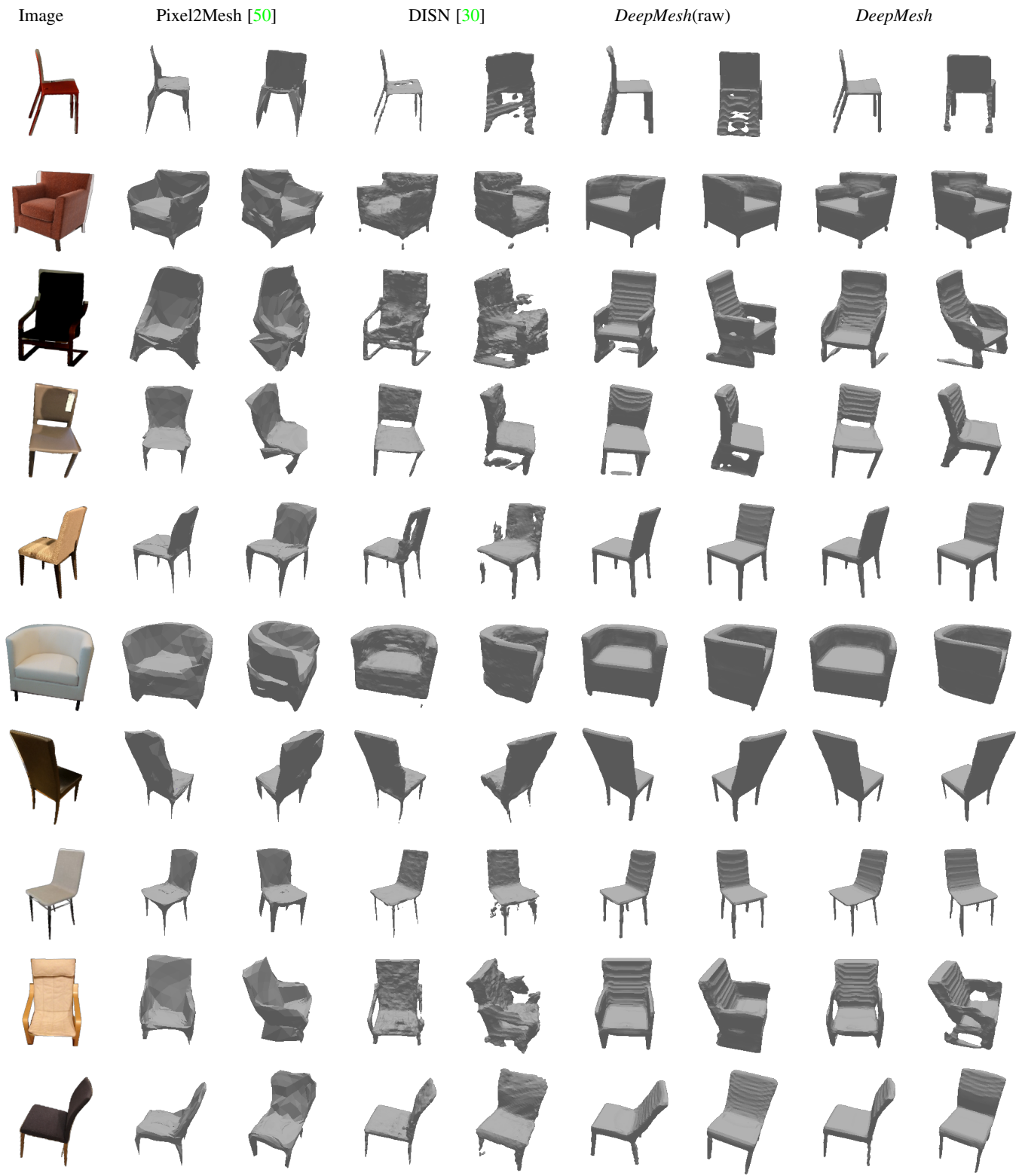
\begin{figure*}[h!]
		\begin{center}
			\begin{overpic}[clip, trim=0.0cm 0cm 0 0.0cm,width=1.0\textwidth]{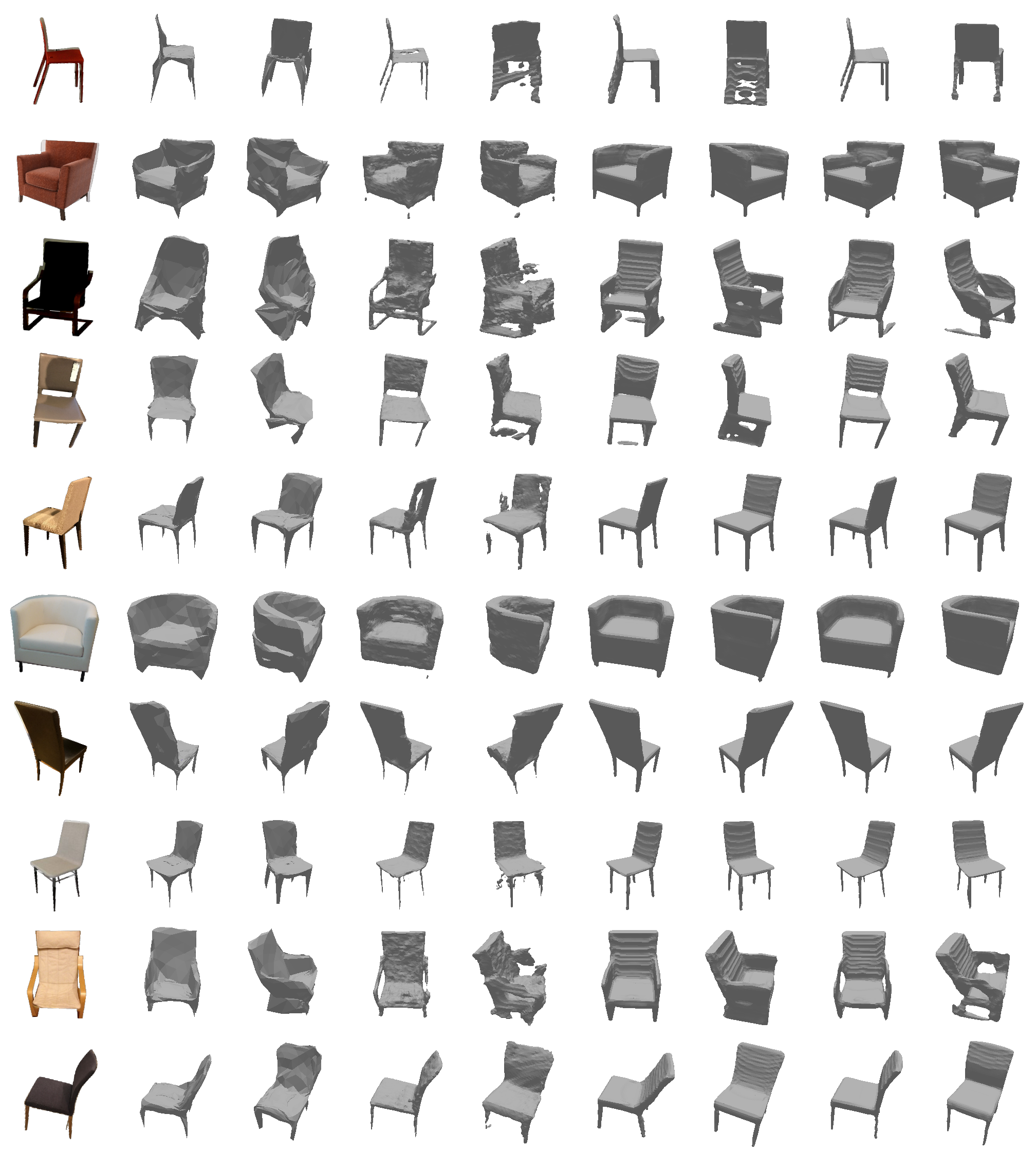}
			\put(2,101){\small{Image }}
			\put(14,101){\small{Pixel2Mesh \cite{Wang18e}}}
		    \put(36,101){\small{DISN \cite{Xu19b}}}
			\put(54,101){\small{\textit{DeepMesh}(raw)}}
			\put(75,101){\small{\textit{DeepMesh}}}
			\end{overpic}
		\end{center}
		\vspace{-10pt}
		\caption{\textbf{Comparative results for SVR on Pix3D.} We compare our refined predictions to  runner-up approaches for the experiment in Tab. \ref{tab:pix3d}. \textit{DeepMesh} can represent arbitrary topology as well as learn strong shape priors, resulting in reconstructions that are consistent even when observed from view-points different from the input one. For more results see Appendix.}
\label{fig:pix3D}
\vspace{-6pt}
\end{figure*}

We report our results in Tab.~\ref{tab:pix3d} and in Fig.~\ref{fig:pix3D}. Interestingly, in this more challenging setting using real-world images, our simple baseline \textit{DeepMesh} (raw) already performs on par with more sophisticated methods that use camera information \cite{Xu19b}. As for ShapeNet, our full model \textit{DeepMesh} outperforms all other approaches.


\begin{table}[t]
	\centering
	\small
	\caption{\textbf{SVR comparative results on Pix3D Chairs.} Our full approach outperforms our best competitor in all metrics on real images.}
	\vspace{-4mm}
	\label{tab:pix3d}
	\begin{tabularx}{\linewidth}{@{}X*1{@{\hspace{2mm}}c}@{\hspace{3mm}}c@{\hspace{2mm}}c@{}}
		\toprule
		\myfootnotesize Metric & \myfootnotesize DISN \cite{Xu19b} & \myfootnotesize DeepMesh (raw) & \myfootnotesize DeepMesh\\
		\midrule
		CHD $\cdot 10^3$ $\downarrow$   & 5.150 & 4.850
& \textbf{4.063}($\downarrow$ 16.3$\%$)
  \\
	     NC $\%$ $\uparrow$ & 56.94  & 62.76  &  \textbf{64.28} ($\uparrow$ 2.4 $\%$) \\
		\bottomrule
	\end{tabularx}
\end{table}

\subsection{Shape Optimization}
\label{sec:cfd}

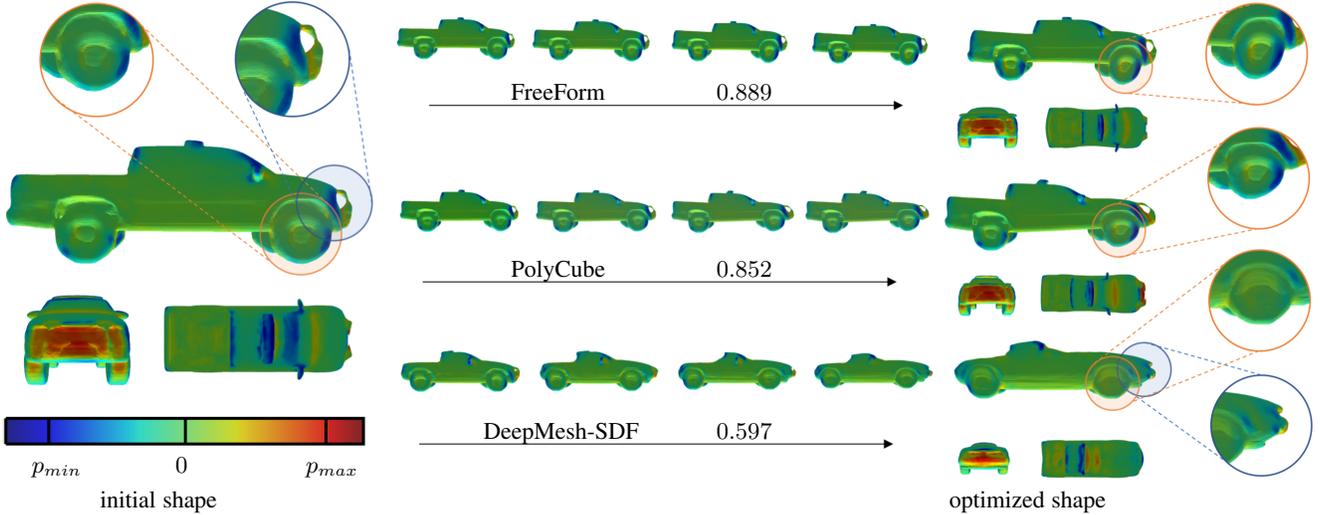
\begin{figure*}[t]
		\begin{center}
			\begin{overpic}[clip, trim=2.5cm 3cm 0cm 4cm,      
			                        width=\textwidth]{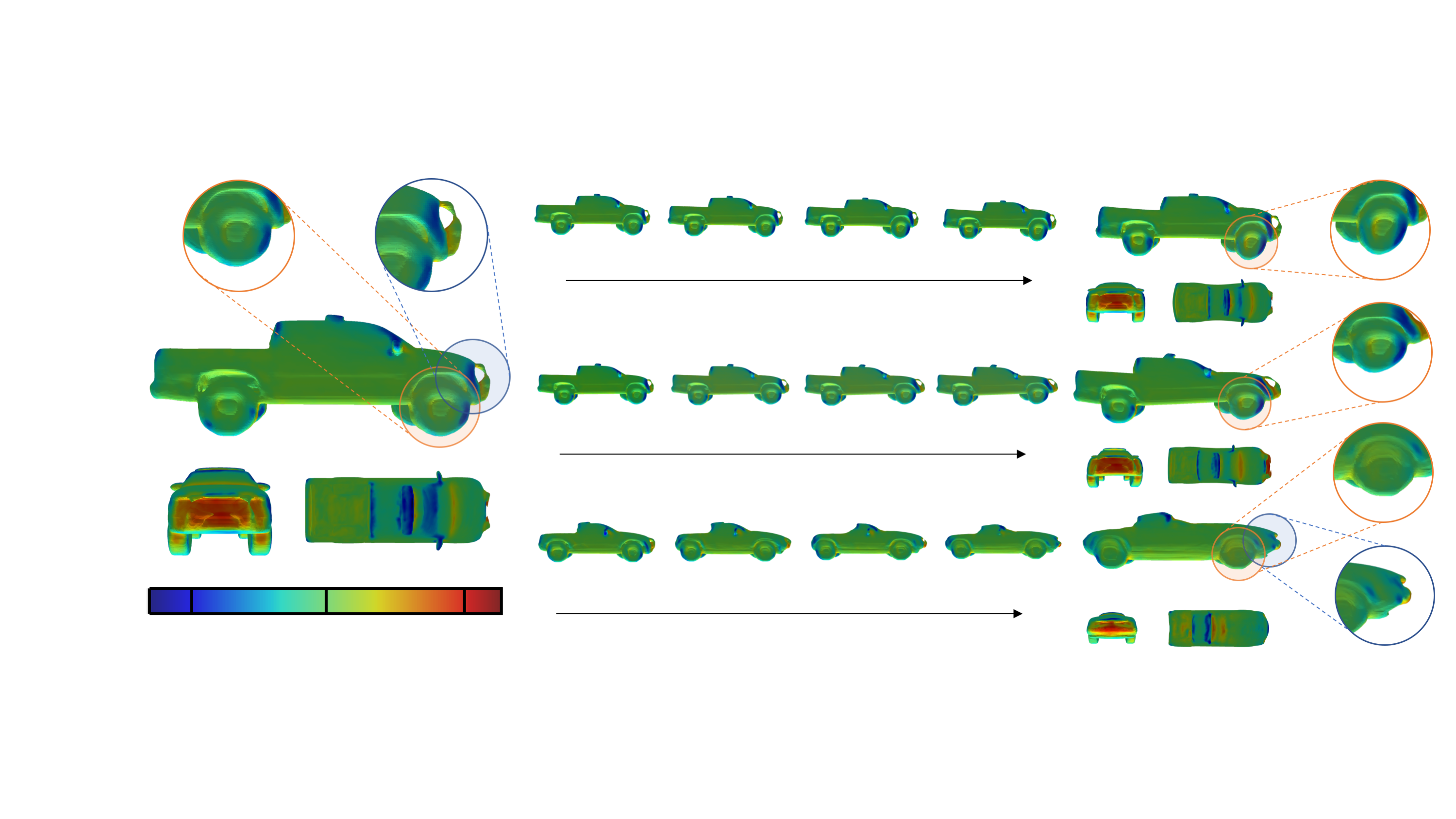}
    			\put(38,6){\small{{DeepMesh-SDF}}}
    			\put(40,17.83){\small{PolyCube }}
    			\put(40,30.75){\small{FreeForm }}
    			
    			\put(25, 3.5){\small{$p_{max}$}}
    			\put(15.5, 3.5){\small{$0$}}
    			\put(5, 3.5){\small{$p_{min}$}}

    			\put(55,6){\small{$0.597$}}
    			\put(55,17.83){\small{$0.852$}}
    			\put(55,30.75){\small{$0.889$}}
			
			\put(72, 01){\small{optimized shape}}
    			\put(10, 01){\small{initial shape}}
    			
			\end{overpic}
		\end{center}
\vspace{-4mm}
\caption{\textbf{Drag minimization.} Starting from an initial shape (left column), $\mathcal L _{\textit{task}}$ is minimized  using three different parameterizations: FreeForm (top), PolyCube (middle), and our \textit{DeepMesh} (bottom). The middle column depicts the optimization process and the relative improvements in terms of $\mathcal L _{\textit{task}}$. The final result is shown in the right column.
FreeForm and PolyCube lack a semantic prior, resulting in implausible details such as sheared wheels (orange inset). By contrast, \textit{DeepMesh} not only enforces such priors but can also effect topology changes (blue inset).}
\label{fig:cfd_three_baselines}
\end{figure*}

\begin{table*}[h]
	\centering
	\small
	\caption{\textbf{CFD-driven optimization}.We minimize drag on car shapes comparing different surface parameterizations. Numbers in the table (avg $\pm$ std) denote {relative improvement of the objective function $\mathcal L _{\text{task}}^{\%} = \mathcal L _{\text{task}} / \mathcal L _{\text{task}}^{t=0} $ } for the optimized shape, as obtained by CFD simulation in OpenFoam.}
	\label{tab:cfd_results}
	\begin{tabularx}{\linewidth}{@{}X*4{@{\hspace{2mm}}c}@{\hspace{3mm}}c@{\hspace{2mm}}c@{}}
		\toprule
		\myfootnotesize Parameterization & \myfootnotesize None & \myfootnotesize Scaling & \myfootnotesize FreeForm \cite{Baque18}  & \myfootnotesize PolyCube \cite{Umetani18} & \myfootnotesize DeepMesh-SDF & \myfootnotesize DeepMesh-OCC\\
		\midrule
		Degrees of Freedom  & $\sim100$k &  $3$  & $21$  &  $\sim 332$ & $256$ & $256$\\
		Simulated {$\mathcal L _{\text{task}}^{\%} $} $\downarrow$ 
		& not converged  & $0.931 \pm 0.014$ &  $0.844 \pm 0.171$ & $0.841 \pm 0.203$ & $\textbf{0.675} \pm \textbf{0.167}$ & $0.721 \pm 0.154$\\
		\bottomrule
	\end{tabularx}
\end{table*}

Computational Fluid Dynamics (CFD)  plays a central role in designing cars, airplanes and many other machines. It typically involves approximating the solution of the Navier-Stokes equations using numerical methods. Because this is computationally demanding, \textit{surrogate} methods~\cite{Toal11,Xu17,Baque18,Umetani18} have been developed to infer physically relevant quantities, such as pressure fields, drag, and lift directly from 3D surface meshes without performing actual physical simulations. This makes it possible to optimize these quantities with respect to the 3D shape using gradient-based methods and at a much lower computational cost. 

In practice, the space of all possible shapes is immense. Therefore, for the optimization to work well, one has to parameterize the space of possible shape deformations, which acts as a strong regularizer. In~\cite{Baque18,Umetani18} hand-crafted surface parameterizations were introduced. It was effective but not generic and had the potential to significantly restrict the space of possible designs. We show here that we can use \textit{DeepMesh} to improve upon hand-crafted parameterizations.

\subsubsection{Experimental Setup.}

We started with the ShapeNet car split by automatic deletion of all the internal car parts \cite{Sin13} and then manually selected $N=1400$ shapes suitable for CFD simulation. For each surface ${\mathcal M _i}$ we ran OpenFoam~\cite{Jasak07} to predict a pressure field $p_i$ exerted by air traveling at 15 meters per second towards the car. The resulting training set  $\{ \mathcal M _i, p_i \}_{i=1}^N$ was then used to train a Mesh Convolutional Neural Network~\cite{Fey18} $g_{\beta}$ to predict the pressure field $p=g_{\beta}(\mathcal M)$, as in~\cite{Baque18}. We use $\{\mathcal M_i\}_{i=1}^N$ to also learn the representation of Sec.~\ref{sec:diff} and train the network that implements $f_{\Theta}$ of Eq.~\ref{eq:imp}. {As in Section~\ref{sec:svr}, we train both an occupancy network and signed-distance  network, which we dub \textit{DeepMesh-OCC} and \textit{DeepMesh-SDF}, respectively.}

The shapes are deformed to minimize the aerodynamic objective function
\begin{align}
\label{eq:cfd}
    \mathcal L _{\text{task5}}(\mathcal{M}) = \iint _{\mathcal{M}} \, g_\beta \, \mathbf n_x \, \text{d}\mathcal{M} + \mathcal L _{\text{constraint}}(\mathcal{M}) +  \mathcal L _{\text{reg}}(\mathcal{M})\; ,
\end{align}
where  $\mathbf n_x$ denotes the projection of surface normals along airflow direction, the integral term approximates drag given the predicted pressure field \cite{Munson13}, $\mathcal{L}_{\text{constraint}}$ is a loss that forces the result to preserve space the engine and the passenger compartment, and  $\mathcal{L}_{\text{reg}}$ is a regularization term that prevents $\bz$ from moving too far away from known shapes.  $\mathcal{L}_{\text{constraint}}$ and $\mathcal{L}_{\text{reg}}$ are described in more detail in the supplementary material.

\subsubsection{Comparative Results.}

We compare our surface parameterizations to several baselines: (1) vertex-wise optimization, that is, optimizing the objective with respect to each vertex; (2) scaling the surface along its 3 principal axis; (3) using the \textit{FreeForm} parameterization of~\cite{Baque18}, which extends scaling to higher order terms as well as periodical ones and (4) the \textit{PolyCube} parameterization of~\cite{Umetani18} that deforms a 3D surface by moving a pre-defined set of control points. 

We report quantitative results for the minimization of the objective function of Eq.~\ref{eq:cfd} for a subset of 8 randomly chosen cars in Table~\ref{tab:cfd_results}, and show qualitative ones in Fig.~\ref{fig:cfd_three_baselines}. Not only does our method deliver lower drag values than the others but, unlike them, it allows for topology changes and produces semantically correct surfaces as shown in Fig.~\ref{fig:cfd_three_baselines}(c). We provide additional results in the supplementary material. As can be seen in Table~\ref{tab:svr_ablation}, \textit{DeepMesh-SDF} slightly outperforms \textit{DeepMesh-OCC}. We conjecture this is due to our sampling strategy, which follows closely the one of \cite{Park19c} {and might therefore favor SDF networks}.

\subsection{Scene Reconstruction}
\label{sec:reconstruction}
\begin{table}[t]
  \centering
  \caption{\textbf{Refining reconstructed scenes from ConvOccNet}. }
  \label{table:convoccnet}
  \begin{tabular}{l|cc}
      & Original & Optimized \\ \hline
  CHD &    0.954     & \textbf{0.529}         \\
  IoU &  81.67 \%      & \textbf{88.74 \%}        
  \end{tabular}
\end{table}


\begin{figure}[t]
		\hspace{-2pt}
		\begin{center}
			\begin{overpic}[width=0.49\textwidth]{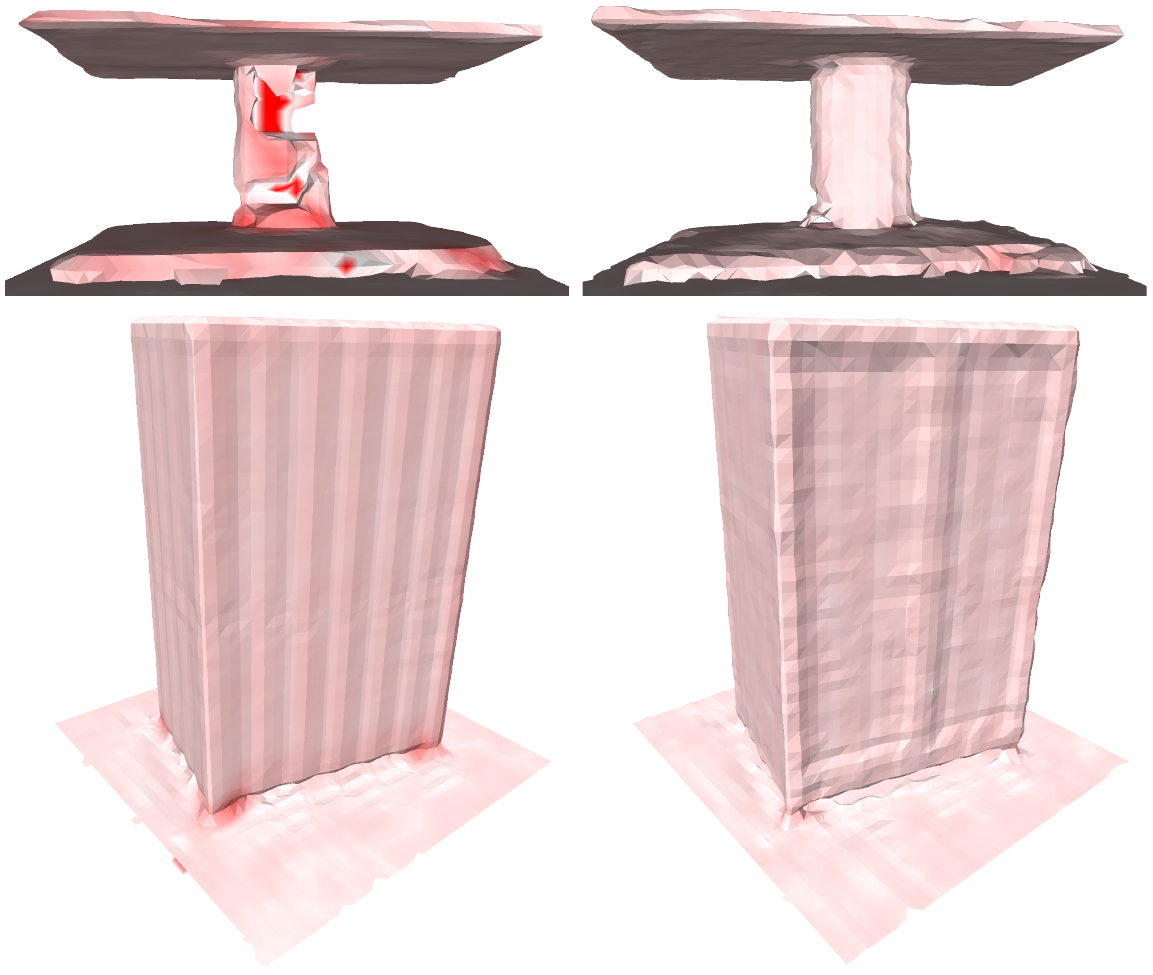}
				\put(20,-4){\small{{Raw}}}
				\put(70,-4){\small{{Refined}}}
			\end{overpic}
		\end{center}
		\caption{\textbf{Refining CON~\cite{Peng20c} scenes.} 
		Using our differentiable surface extraction, we can refine CON features so that the output mesh better matches the input scene point cloud. This fixes artefacts on the table, and reconstructs finer details on the wardrobe from the scene in Fig.~\ref{fig:teaser}(c - left). Chamfer error are shown in red, and reduced after refinement.
		}
\label{fig:CON_crop}
\vspace{-6pt}
\end{figure}

In the examples of Section~\ref{sec:svr} and~\ref{sec:cfd}, we had access to code and training data that enabled us to compare the performance of SDFs and occupancy grids and found out experimentally that the former tend to perform better. However, there are situations in which we only have access to a network that produces occupancy fields without any easy way to transform it into one that produces SDFs. In this section, we show that the ability of our method to handle not only SDFs but also occupancy fields is valuable.

We use the pretrained scene reconstruction network of~\cite{Peng20c} that regresses an occupancy field from sparse point clouds describing indoor scenes. We use 10k points as in the original paper. A point cloud $P$ is first encoded into a set of feature vectors of size 32. These are stored over a 3D feature grid $G \in  \mathbb{R}^{32\times 32\times 32\times 32}$ and in three {projected} 2D feature planes $P_1, P_2, P_3 \in \mathbb{R}^{128\times 128\times 32}$, all aligned with the input point cloud. These features are linearly interpolated in space and decoded into occupancy values. We transform the resulting occupancy field into a differentiable mesh $M_{\bz}$ using our framework, where $\bz = [G | P_1 | P_2 | P_3]$ is the concatenation of the feature grids.

Using the differentiability of the mesh, we minimize with respect to $\bz$ the single-sided Chamfer distance
\begin{equation}
\mathcal{L}_{\text{task6}} = \sum_{p \in P} \min_{a \in M_{\bz}} \left \| a - b \right \|^2  \; ,
\end{equation}
between the reconstructed mesh and $P$ , where $a \in M_{\bz}$ represents 10k points sampled over the mesh.

In Tab.~\ref{table:convoccnet} we compute the average Chamfer distance and Intersection over Union (IoU)~\cite{Peng20c} with the ground truth meshes for the 2 provided test scenes. The improvements are substantial, as can be seen in Fig.~\ref{fig:teaser}(c) and Fig.~\ref{fig:CON_crop}.

\subsection{End-to-End Training}

\begin{table}[t]
		\centering
		\small
		\caption{\textbf{End-to-end training.} We exploit end-to-end-differentiability to fine-tune pre-trained \textit{DeepSDF} networks in order to minimize directly surface-to-surface (Chamfer) distance. This improves Chamfer distance (CHD) and normal consistency (NC) on the testset.
		}
		\label{table:end2end_new}
		\begin{tabularx}{\linewidth}{@{}X*1{@{\hspace{2mm}}|c}@{\hspace{3mm}}c@{\hspace{2mm}}c@{}}
			\toprule
			Metric & Category &  \textit{DeepSDF} &  \textit{DeepMesh} \\
			\midrule
			\multirow{2}{*}{CHD $\cdot 10^4$ ($\downarrow$)} 	& \textit{Chairs}  &  2.75  & \textbf{2.56} ($\downarrow$ 6.9 $\%$) \\
																										 		& \textit{Lamps}  &  8.12  & \textbf{7.59}  ($\downarrow$ 6.5 $\%$)\\
			\midrule
			\multirow{2}{*}{NC ($\uparrow$)} & \textit{Chairs}  &  80.9  & \textbf{81.9} ($\uparrow$ 1.2 $\%$)\\
																			 & \textit{Lamps}  &  73.9  & \textbf{75.1} ($\uparrow$ 1.6 $\%$)\\
			\bottomrule
		\end{tabularx}
\end{table}


\begin{figure}[t]
		\hspace{-2pt}
		\begin{center}
			\begin{overpic}[width=0.49\textwidth]{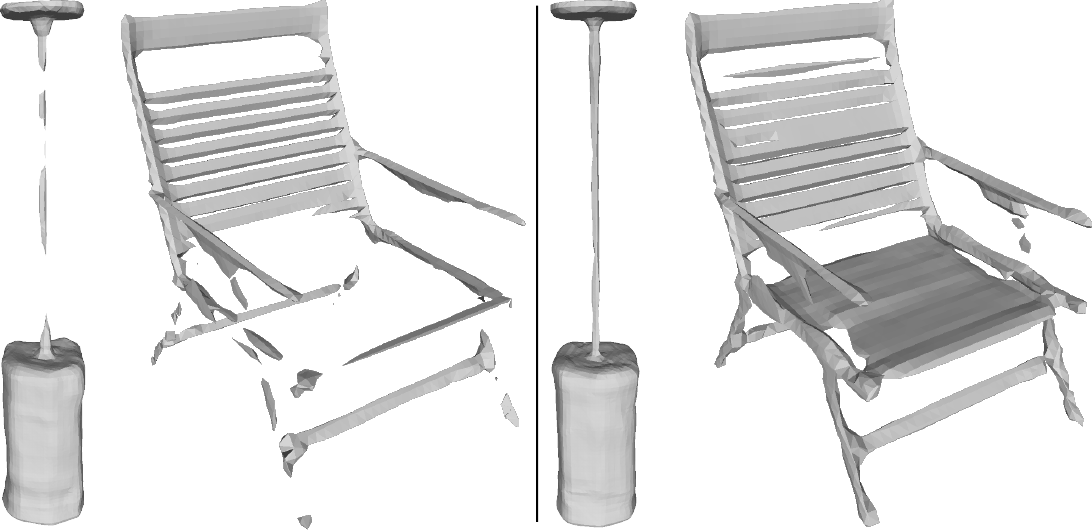}
				\put(18,-4){\small{\textit{DeepSDF}}}
				\put(68,-4){\small{\textit{DeepMesh}}}
			\end{overpic}
		\end{center}
		\caption{\textbf{Learning network weights by minimizing the Chamfer loss.}
		\textbf{DeepSDF} is trained by minimizing errors on predicted SDF values, which can result in thin components being poorly reconstructed. \textbf{DeepMesh} is trained by also minimizing Chamfer distances, which penalizes such mistakes. The resulting network reconstructs thin structures better.
		}
\label{fig:finetuning}
\vspace{-6pt}
\end{figure}

In all previous examples, we considered a pre-trained network $f_{\Theta}$ and optimized with respect to the latent variables it takes as input. We now demonstrate that our differentiable isosurface extraction scheme can also be used to train  $f_{\Theta}$ and to backpropagate gradients to the network weights $\Theta$. In other words, we consider the setting where $\bc=\Theta$ in Theorem~\ref{th:shapeDerivative} and {show how it can be used to improve the performance of a \textit{DeepSDF} network. } 

Let us therefore assume that the network $f_{\Theta}$ implements \textit{DeepSDF}, as described in~\cite{Park19c}. In the original paper, $\Theta$ along with the latent representations are learned by minimizing the implicit loss function  $\mathcal L _{\text{imp}}$ of Eq.~\ref{eq:imp} and its accuracy is assessed in terms of the chamfer distance between target shapes and reconstructed ones. It can be written as 
\begin{align}
\label{eq:cham_finetuning}
\mathcal L_{\text{chamfer}} = \sum_{\mathbf p \in P} \min_{\mathbf q \in Q} \| \mathbf p - \mathbf q \| _2 ^ 2 + \sum_{\mathbf q \in Q} \min_{\mathbf p \in P} \| \mathbf p - \mathbf q \| _2 ^ 2 \;  ,
\end{align}
where $P$ and $Q$ denote surface samples, 10K in our implementation. Computing $L_{\text{chamfer}}$ requires triangulating, which can be done differentiably in our framework.  This gives us the option to train $f_{\Theta}$ not only by minimizing $L_{\text{chamfer}}$, as in the original method, but by minimizing $\mathcal L _{\text{imp}} + \mathcal{L}_{\text{chamfer}}$. In other words, we can optimize directly with respect to a relevant metric.

In practice, we first minimize $\mathcal L _{\text{imp}}$ to learn a first version of $\Theta$ and of the latent vectors $\bz$ of Eq.~\ref{eq:imp}. As described in~\cite{Park19c}, this yields a network that we will refer to as \textit{DeepSDF}. We then freeze the latent vectors and minimize $\mathcal L _{\text{imp}} + \mathcal{L}_{\text{chamfer}}$ with respect to $\Theta$. This yields a second network that we will refer to as \textit{DeepMesh}. We do this for the \textit{chairs} and \textit{lamps} categories of ShapeNet~\cite{Chang15}. For chairs, we use the same data split and samples as in Sec.~\ref{sec:svr}. We apply the same pre-processing steps to lamps, remove duplicates from the original dataset, and use 1100 training shapes and 106 testing shapes.

We compare  \textit{DeepSDF} and  \textit{DeepMesh} qualitatively in  Fig.~\ref{fig:finetuning} and quantitatively in Tab.~\ref{table:end2end_new}, where we report metrics on the test sets by fitting latent codes to SDF samples of unseen shapes. Minimizing $\mathcal L_{\text{chamfer}}$ delivers a substantial boost. This is especially true for lamps because they feature thin structures for which even a small error in the predicted SDF values can result in a substantial surface misalignment. 

\subsection{Execution Speed}

\begin{table}[t]
\centering
\caption{\textbf{Execution speed} for forward and backward passes, either directly in the implicit domain or with our method providing an explicit mesh. For using a mesh, we list runtimes of a naive and an optimized implementation of isosurface extraction.}
\label{tab:execution_speed}
\vspace{-2mm}
	\begin{tabularx}{0.48\textwidth}{lXc}
	\Xhline{\lightrulewidth}\addlinespace[0.4em]
	\multicolumn{3}{l}{\textbf{In the implicit domain}} \\
	Operation & & Time \\ \hline
	Forward: 	& \textit{compute fields values for $8192$ points}	& 4 ms.\\
	Backward:	& \textit{from fields values to latent code}   	& 8 ms.\\

	\Xhline{\lightrulewidth}\addlinespace[0.4em] 
	\multicolumn{3}{l}{\textbf{With an explicit mesh}}  \\
	Operation & & Time \\ \hline
	Forward, naive: 		& \textit{create mesh with dense grid + CPU marching cubes}	& 785 ms.\\
	Forward, optimized: & \textit{create mesh with sparse grid + GPU marching cubes}	& 29 ms.\\
	Backward: 					& \textit{from mesh vertices to latent code}                  				& 6 ms.
	\end{tabularx}
\end{table}

We now turn to measuring the execution speed of our method and the overhead it incurs over a simple supervision of implicit fields values. In Tab.~\ref{tab:execution_speed}, we compare forward and backward times for losses either on the field's values ($\mathcal{L}_{data}$ of Eq.~\ref{eq:sdf}) or through isosurface extraction ($\mathcal L_{\text{chamfer}}$ of Eq.~\ref{eq:cham_finetuning}). The network is a DeepSDF with 8 layers of size 512, and we report average times over the testing chairs of ShapeNet. For $\mathcal{L}_{data}$ we apply it on the default amout of $8192$ points per batch. For $\mathcal L_{\text{chamfer}}$ we run isosurface extraction at resolution $128^3$. The machine we run this test on uses an NVidia V100 GPU with an Intel Xeon Gold 6240 CPU.

A naive isosurface extraction is 2 orders of magnitude slower than simply computing SDF values. However, with the coarse-to-fine strategy presented in Sec.~\ref{subsec:forward_backward}, the overhead is reasonable and allows for efficient training. Note also that the backwards pass of our method is slightly faster than with direct supervision of SDF values. This is because we backpropagate from the surface points only, instead of samples over the entire volume.



\section{Conclusion}

We have introduced \textit{DeepMesh}, a new approach to extracting 3D surface meshes from continuous deep implicit fields while preserving end-to-end differentiability. This makes it possible to combine powerful implicit models with objective functions requiring explicit representations such as surface meshes. 

\textit{DeepMesh} has the potential to become a powerful Computer Assisted Design tool because allowing differential topology changes of explicit surface parameterizations opens the door to new applications. In future work, we will further extend our paradigm to Unsigned Distance Functions to handle open surfaces without having to thicken them, as we did here. We also plan to exploit Generative Adversarial Networks operating on surface meshes~\cite{Cheng19b} to increase the level of realism of the surfaces we generate. Furthermore, our method still requires 3D supervision on the field at training time. In the future, we plan to address this with recent approaches that allow learning implicit representations from raw data~\cite{Atzmon20}.



\ifCLASSOPTIONcompsoc
  \section*{Acknowledgments}
\else
  \section*{Acknowledgment}
\fi

This work was supported in part by the Swiss National Science Foundation.

\bibliographystyle{IEEEtran}
\bibliography{string,vision,learning,cfd,graphics,optim,biomed,misc,local}

\clearpage
\section{Supplementary Material}

In this supplementary material, we first remind the reader of why Marching Cubes is not differentiable. 
We also provide additional details about our experiments on single view 3D reconstruction and drag minimization.


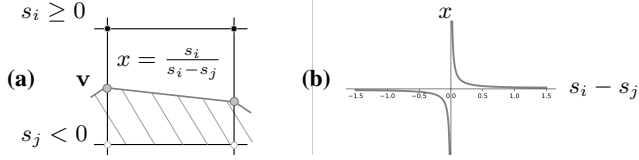
\begin{figure}[ht]
\vspace{-6pt}
\hspace{6pt}
            \begin{center}
			\begin{overpic}[clip, trim=0.0cm 10cm 12cm 4cm,width=0.5\textwidth]{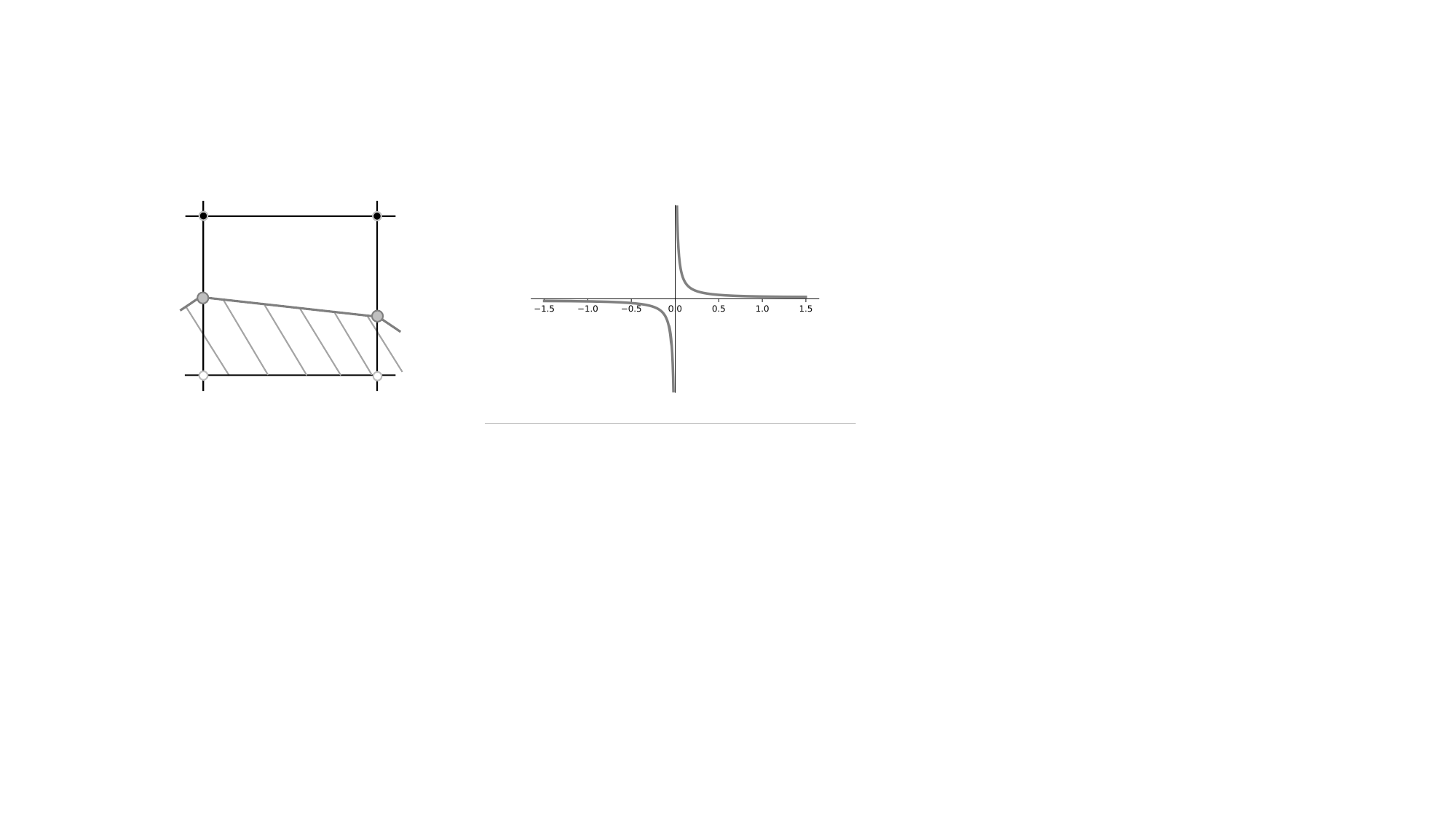}
			\put(9,20){\small{$ s _i \geq 0 $}}
			\put(9, 2){\small{$ s _j < 0 $}}
			\put(17,10){\small{$ \mathbf v $}}
			\put(23,13){\small{$ x = \frac{s _i}{s _i - s _j } $}}

			\put(7,10){\small{\textbf{(a)}}}
			\put(50,10){\small{\textbf{(b)}}}
			\put(89,9){\small{$s_i - s_j$}}
			\put(70,20){\small{$x$}}

			\end{overpic}

			\end{center}
	
	\caption{\textbf{Marching cubes differentiation.} \textbf{(a)} Marching Cubes determines the relative position $x$ of a vertex $\bv$ along an edge via linear interpolation. This does not allow for effective back-propagation when topology changes because of a singularity when $s _i=s _j$. \textbf{(b)} We plot $x$, relative vertex position along an edge. Note the infinite discontinuity for $s _i=s _j$. }
		\label{fig:mc_supp}
\end{figure}

\subsection{Non-differentiability of Marching Cubes}

The Marching Cubes (MC) algorithm \cite{Lorensen87} extracts the zero level set of an implicit field and represents it \textit{explicitly} as a set of triangles. As discussed in the related work section, it comprises the following steps: (1) sampling the implicit field on a discrete 3D grid, (2) detecting zero-crossing of the field along grid edges, (3) assembling surface topology, that is,  the number of triangles within each cell and how they are connected, using a lookup table and (4) estimating the vertex location of each triangle by performing linear interpolation on the sampled implicit field. These steps can be understood as topology estimation followed by determination of surface geometry.

More formally, let $S = \{ s_i \} \in \mathbb R ^ {N \times N \times N}$ be an implicit field sampled over a discrete Euclidean grid $G_{3D} \in \mathbb R ^ {N \times N \times N \times 3}$, where $N$ denotes the resolution along each dimension. Within each voxel, surface topology is determined based on the sign of $s_{i}$ at its 8 corners. This yields $2^8 = 256$ possible surface topologies within each voxel. Once they have been assembled into a consistent surface, vertices are created when the implicit field changes sign along one of the edges of the voxel.
In such cases, the vertex location $\mathbf v$ is determined by linear interpolation. Let $x \in [0,1]$ denote the vertex relative location along an edge $(\mathbf G_i, \mathbf G_j)$, where $\mathbf G_i$ and $\mathbf G_j$ are grid corners such that $s_j < 0$ and $s_i \geq 0$. This implies that, if $x = 0$, then $\mathbf v = \mathbf G_i$ and conversely if  $x = 1$ then $\mathbf v = \mathbf G_j$. In the MC algorithm, $x$ is is determined as the zero crossing of the interpolant of $s_i$, $s_j$, that is, 
\begin{align}
    x = \frac{s_i}{s_i - s_j} \; ,
\end{align}
as shown in Fig. \ref{fig:mc_supp}(a). The vertex location is then taken to be
\begin{align}
    \mathbf v = \mathbf G_i + x (\mathbf G_j - \mathbf G_i).
\end{align}
Unfortunately, this function is discontinuous for $s_i = s_j$, as illustrated in Fig \ref{fig:mc_supp}(b). Because of this, we cannot swap the signs of $s_i$ and $s_j$ during backpropagation. This prevents topology changes while differentiating, as discussed in~\cite{Liao18a}.

\subsection{Modeling non Watertight Surfaces}

\begin{figure*}[t]

            \begin{center}
			\begin{overpic}[clip, trim=0.0cm 7cm 3cm 0cm,width=1.1\textwidth]{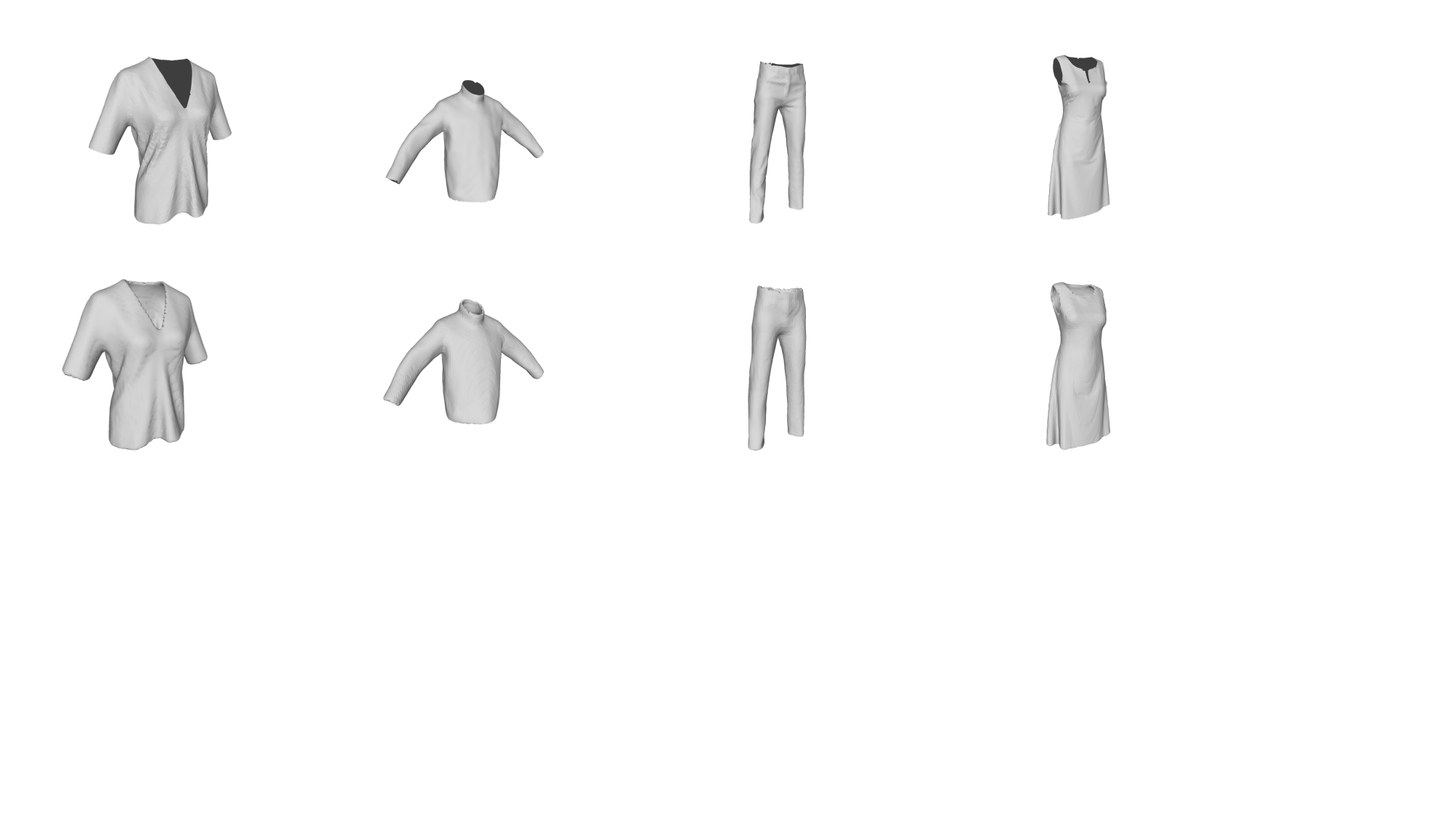}
			\put(0,24){\rotatebox{90}{\small{ground truth}}}
			\put(0, 7){\rotatebox{90}{\small{reconstruction}}}

			\end{overpic} 
			\end{center}
	\vspace{-7mm}
	\caption{\textbf{Representing garments.} Our differentiable mesh parameterization can also be used to represent non-watertight surfaces. Here, we show reconstructions for the experiment presented in Section 4.1. of the main manuscript as well as ground truth open surfaces.  
	}
		\label{fig:garments}
\end{figure*}

As explained in the main manuscript, our pipeline can also be used to represent non watertight surfaces, such as garments. We showed reconstructions of pants and t-shirts. In Fig.~\ref{fig:garments}, we provide additional reconstructions. 


\subsection{Meshing an occupancy field}

\begin{figure}[t]
            \begin{center}
			\begin{overpic}[clip, trim=0.cm 0.cm 0cm 0cm,width=0.5\textwidth]{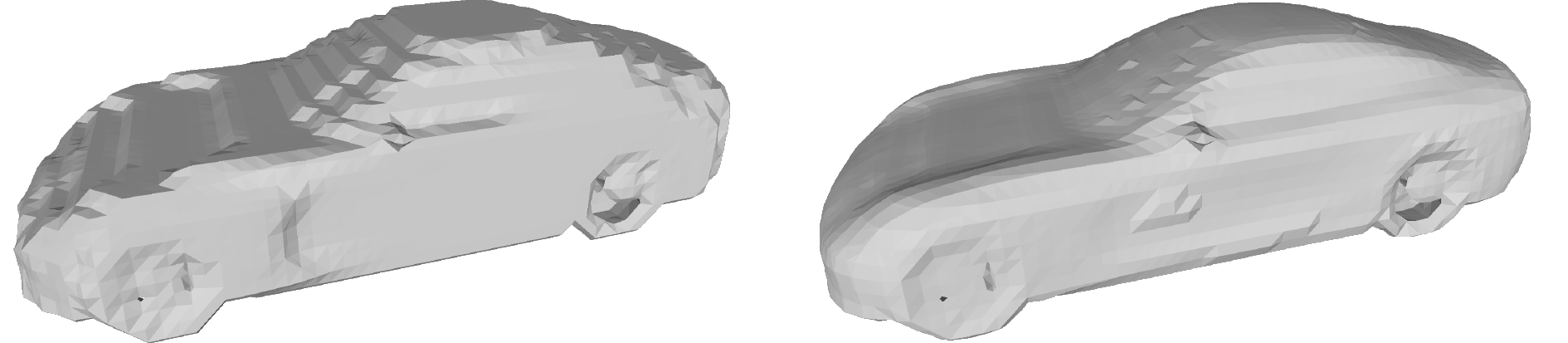}
				\put(25,0){(a)}
				\put(75,0){(b)}
			\end{overpic} 
			\end{center}
	\caption{{\textbf{Marching Cubes on an occupancy field.} \textbf{(a)} running Marching Cubes on the occupancy field predicted by $f_{\Theta}$ yields artifacts. \textbf{(b)}  Using an inverse sigmoid function to amplify the predicted occupancy values yields smoother shapes.}}
	\label{fig:occ_smoothing}
\end{figure}

{
As shown in Fig.~\ref{fig:occ_smoothing}(a), marching cubes is not well-suited for meshing occupancy fields. Its linear interpolation step is designed to approximate vertex locations in case the sampled field is a signed distance function and does not perform well when predicted values are close to binary. Previous work applies mesh smoothing in a post-processing step to mitigate this~\cite{Mescheder19}. We empirically found that amplifying the predicted occupancy values provides a simple but efficient approximation of the linear regime of  a signed distance function. As shown in Fig.~\ref{fig:occ_smoothing}(b), applying an inverse sigmoid function within the Marching Cubes sampling loop yields smoother shapes at a negligible cost. Hence, this is what we do when meshing occupancy fields.
}

\subsection{Comparing against Deep Marching Cubes}

\begin{figure*}[h!]
            \begin{center}
			\begin{overpic}[clip, trim=-2.0cm 14cm 5cm 1cm,width=\textwidth]{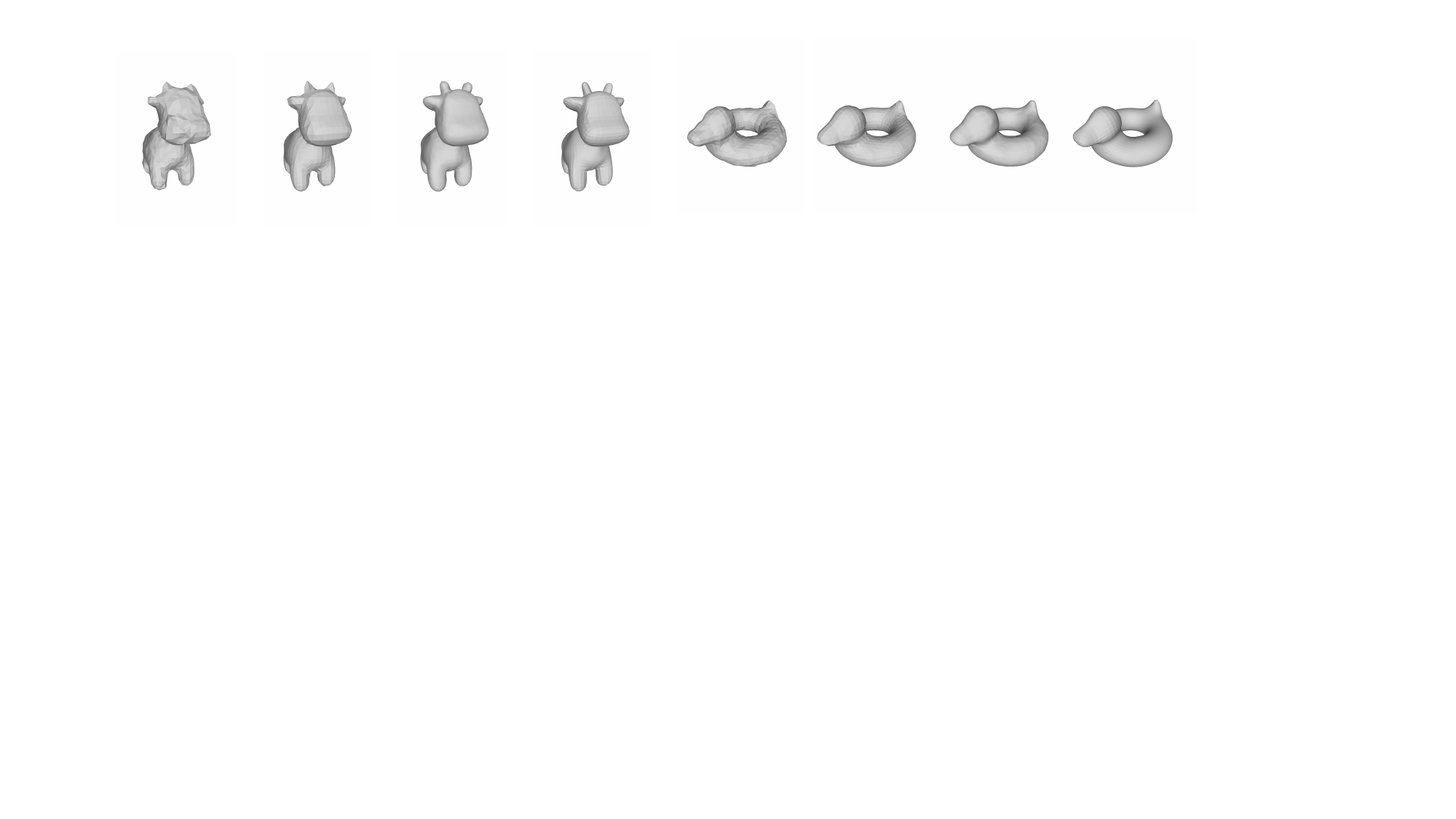}

		    \put(4.5,0){\textbf{\small{CHD$ \cdot 10^2 (\downarrow)$ : }}}
		    
			\put(15.5,12){\small{DMC@$32^3$}}
			\put(17.5,0){\small{$1.87$}}
			\put(26.6,12){\small{Ours@$32^3$}}
			\put(28.5,0){\small{$1.84$}}
			\put(36.5,12){\small{Ours@256$^3$}}
			\put(38.5,0){{\small{{$\mathbf{1.80}$}}}}
			\put(48.5,12.5){\small{ground}}
			\put(49.5,10.7){\small{truth}}
			
			\put(57.5,12){\small{DMC@$32^3$}}
			\put(60.5,0){\small{$1.98$}}
			\put(67.5,12){\small{Ours@$32^3$}}
			\put(70.5,0){\small{$1.94$}}
			\put(77.5,12){\small{Ours@256$^3$}}
			\put(80.5,0){\small{{$\mathbf{1.90}$}}}
			\put(88.5,12.5){\small{ground}}
			\put(89.5,10.7){\small{truth}}
		
			\end{overpic} 
			\end{center}
\vspace{-4mm}
\caption{\textbf{Comparison to Deep Marching Cubes} considering a latent space of size 2. Our metric is Chamfer distance, evaluated on 5000 samples for unit sphere normalized shapes.}
 	\vspace{-4mm}
		\label{fig:reb_dmc}
\end{figure*}

Deep Marching Cubes (DMC) \cite{Liao18a} is designed to convert point clouds into a surface mesh probability distribution. It can handle topological changes but is limited to low resolution surfaces for the reasons discussed in the related work section. In Fig.~\ref{fig:reb_dmc}, we compare our approach to DMC.  {We fit both representations to the rubber duck/cow dataset we introduced in the experiments section.  We use a latent space of size 2 and report our results in terms of the CHD metric.}  As reported in the original paper, we found DMC to be unable to handle grids larger than $32^3$ because it has to keep track of all possible mesh topologies defined within the grid. By contrast, deep implicit fields are not resolution limited. Hence, they can better capture high frequency details.

\subsection{Single view 3D Reconstruction}

\begin{figure*}[t]
		\begin{center}
			\begin{overpic}[clip, trim=0.0cm 7cm 0 4.5cm,width=1.0\textwidth]{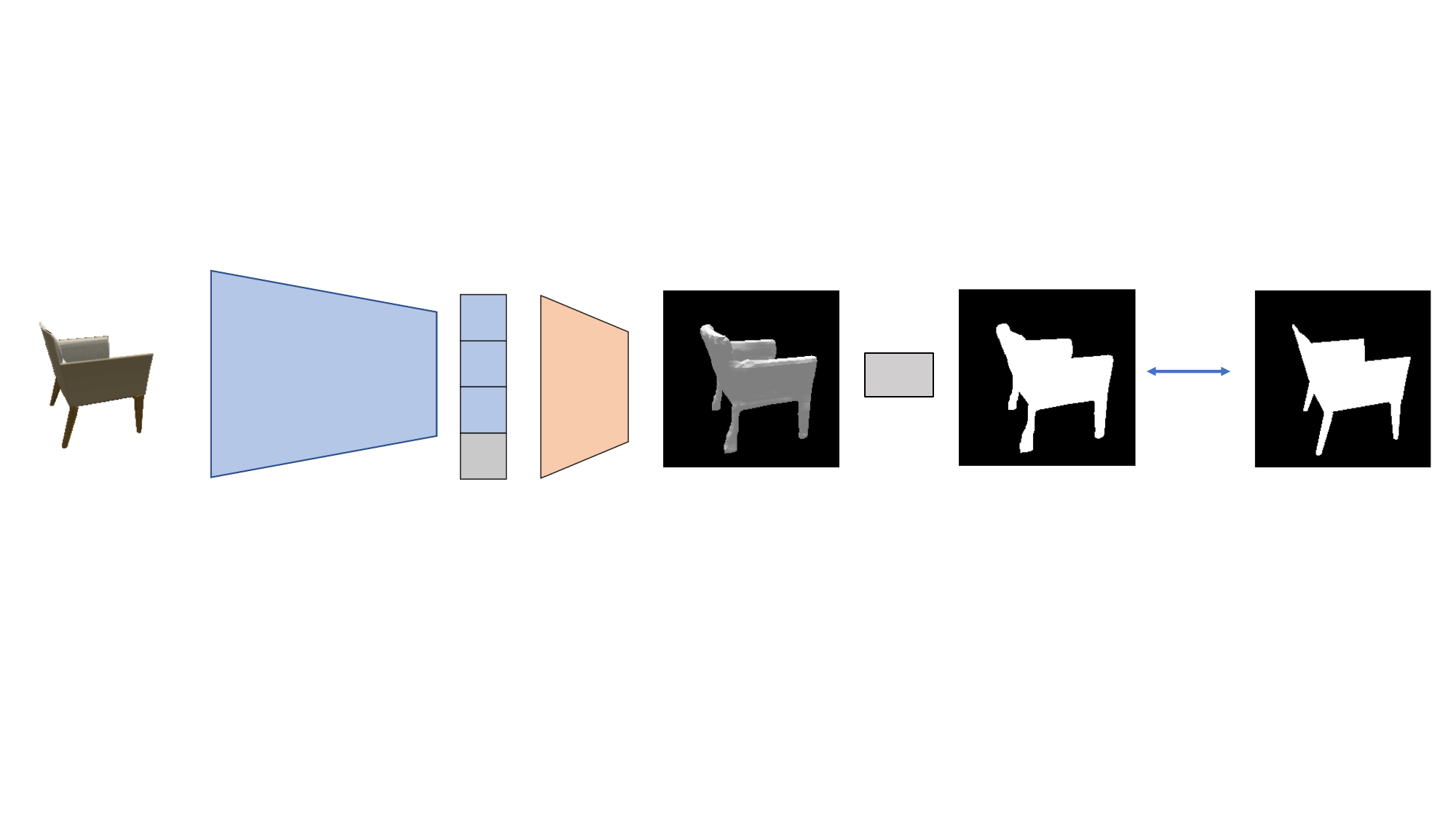}
			\put(4,17.5){\small{Input }}
			\put(18,9){\small{ResNet18}}
			\put(38.1,8.5){\small{MLP}}
			
			\put(32.5,10){\small{$\mathbf z$}}
			\put(32.5,4){\small{$\mathbf x$}}
			
			\put(46.5,17.5){\small{Prediction}}

			\put(59.4,9.3){\small{Reproj}}
			
			\put(67,17.5){\small{Silhouette}}
			
			\put(79.5,11.5){\small{$\mathcal{L}_{\text{task}}$}}
			
			\put(87,18.5){\small{Input}}
			\put(87,16.5){\small{silhouette}}
			
			\end{overpic}
		\end{center}
		\vspace{-9pt}
		\caption{\textbf{Shilouette-driven refinement.} At inference time, given an input image, we exploit the differentiability of \textit{DeepMesh} to refine the predicted surface so that to match input silhouette in image space through Differentiable Rasterization \cite{Kato18} {or contour matching~\cite{Guillard21}}.}
\label{fig:svr_archi}
\vspace{-6pt}
\end{figure*}

We first provide additional details on the Single view 3D Reconstruction pipeline presented in the main manuscript. Then, for each experimental evaluation of the main paper, we first introduce metrics in details, and then provide additional qualitative results. To foster reproducibility, we have made our entire code-base publicly available.

\subsubsection{Architecture}  

Fig \ref{fig:svr_archi} depicts our full pipeline. As in earlier work~\cite{Mescheder19,Chen19c}, we condition our deep implicit field architecture on the input images via a residual image encoder~\cite{He16a}, which maps input images to latent code vectors $\bz$.  Specifically, our encoder consists of a ResNet18 network, where we replace batch-normalization layers with instance normalization ones~\cite{Ulyanov16a} so that to make harder for the network to use color cues to guide reconstruction. These latent codes are then used to condition the signed distance function Multi-Layer Perceptron (MLP) architecture of the main manuscript, consisting of 8 Perceptrons as well as residual connections, similarly to \cite{Park20a}. We train this architecture, which we dub \emph{DeepMesh} (raw), by minimizing $\mathcal L_{\text{imp}}$ (Eq.1 on the main manuscript) wrt. $\Theta$ on a training set of image-surface pairs. 

At inference time, we exploit end-to-end differentiability to refine predictions as depicted in Fig \ref{fig:svr_archi}. 
\comment{That is, given the camera pose associated to the image and the current value of $\bz$, we project vertices and facets into a binary silhouette in image space through a differentiable rasterization function $\text{DR}_{\text{silhouette}}$~\cite{Kato18}. Ideally, the projection matches the observed object silhouette $\cal S$ in the image, which is why we define our objective function as
\begin{equation}
\mathcal{L}_{\text{task}} = \| \text{DR}_{\text{silhouette}}(\mathcal M(\bz))- \mathcal S\|_1 \:, 
\end{equation}
which we minimize with respect to $\bz$. In practice, we run 400 gradient descent iterations using Adam~\cite{Kingma14a} and keep the $\bz$ with the smallest $\mathcal{L}_{\text{task}}$ as our final code vector.}

\subsubsection{Evaluation on ShapeNet.}

Recent work \cite{Tatarchenko19} has pointed out that for a typical shape in the ShapeNet test set, there is a very similar shape in the training set. To mitigate this, we carry our evaluations on new train/test splits which we design by subsampling the original datasets and rejecting shapes that have normal consistency above 98\% for chairs and 96.8\% for cars. Finally, we use the renderings provided in~\cite{Xu19b} for all the comparisons we report.


\begin{figure}[t]
            \begin{center}
			\begin{overpic}[clip, trim=0.5cm 0.5cm 0cm 0cm,width=0.5\textwidth]{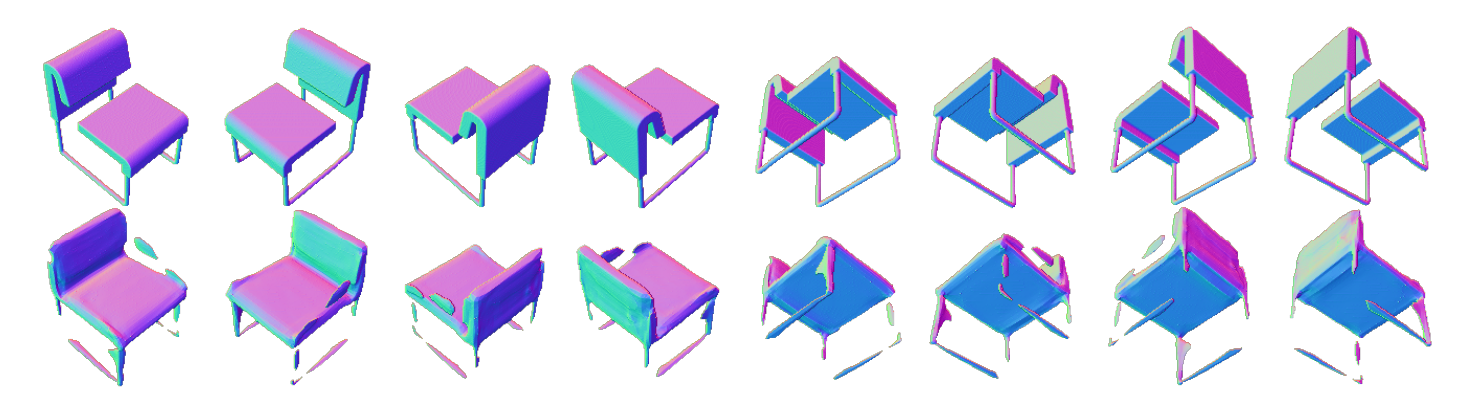}
			
			\end{overpic} 
			\end{center}
	\caption{\textbf{Normal consistency score:} {we render target (top row) and reconstructed shapes (bottom row) normal maps from 8 fixed viewpoints, and take the average of pixelwise cosine similarity to compute an image based normal consistency score.}}
	\label{fig:normal_consist}
\end{figure}

{We use the following SVR metrics for evaluation purposes:}
\begin{itemize}
  \item \textbf{Chamfer $l_2$ pseudo-distance:} Common evaluation metric for measuring the distance between two uniformly sampled clouds of points $P,Q$, defined as 
  \begin{align}
\text{CHD}(P,Q) = \sum_{\mathbf p \in P} \min_{\mathbf q \in Q} \| \mathbf p - \mathbf q \| _2 ^ 2 + \sum_{\mathbf q \in Q} \min_{\mathbf p \in P} \| \mathbf p - \mathbf q \| _2 ^ 2.
\end{align}
We evaluate this metric by sampling $2048$ points from reconstructed and target shape, which are re-scaled to fit into a unit-radius sphere.

  \item \textbf{Normal Consistency:} {We render normal maps for both the target and reconstructed shapes under 8 viewpoints, corresponding to the 8 vertices of a side 2 cube with cameras looking at its center. As depicted on Fig.~\ref{fig:normal_consist}, we get 8 pairs of normal maps that we denote $(\mathbf{n}_i , \widetilde{\mathbf{n}}_i)$ with $1 \leq i \leq 8$. Our normal consistency score is the average over these 8 pairs of images of the pixelwise cosine similarity between the reconstructed and target normal maps. With $\mathbf{n}_i \cap  \widetilde{\mathbf{n}}_i$ being the set of non-background pixel coordinates in both $\mathbf{n}_i$ and $\widetilde{\mathbf{n}}_i$, we have
  \begin{align}
    \text{NC} = \tfrac{1}{8} \sum_{i=1}^{i=8} \tfrac{1}{ |\mathbf{n}_i \cap  \widetilde{\mathbf{n}}_i| } \sum_{(u,v) \in \mathbf{n}_i \cap  \widetilde{\mathbf{n}}_i} \tfrac{\mathbf{n}_i[u,v] \cdot \widetilde{\mathbf{n}}_i[u,v]}{\left \| \mathbf{n}_i[u,v] \right \| \left \| \widetilde{\mathbf{n}}_i[u,v] \right \|}.
    \end{align}
    }

\end{itemize}

\subsubsection{Additional Qualitative Results} 
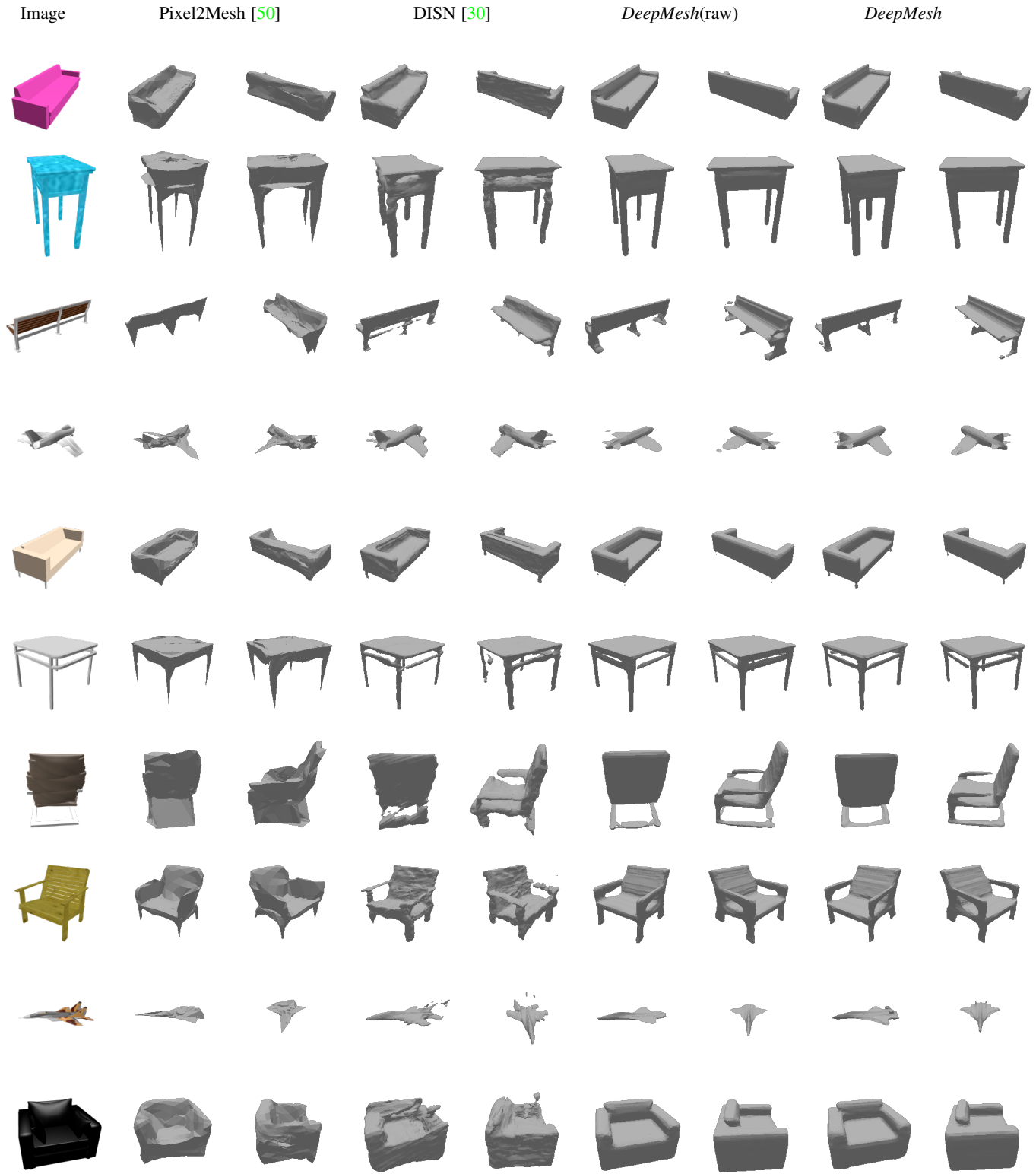
\begin{figure*}[h!]
        \vspace{25pt}
		\begin{center}
			\begin{overpic}[clip, trim=0.0cm 0cm 0 0.0cm,width=1.0\textwidth]{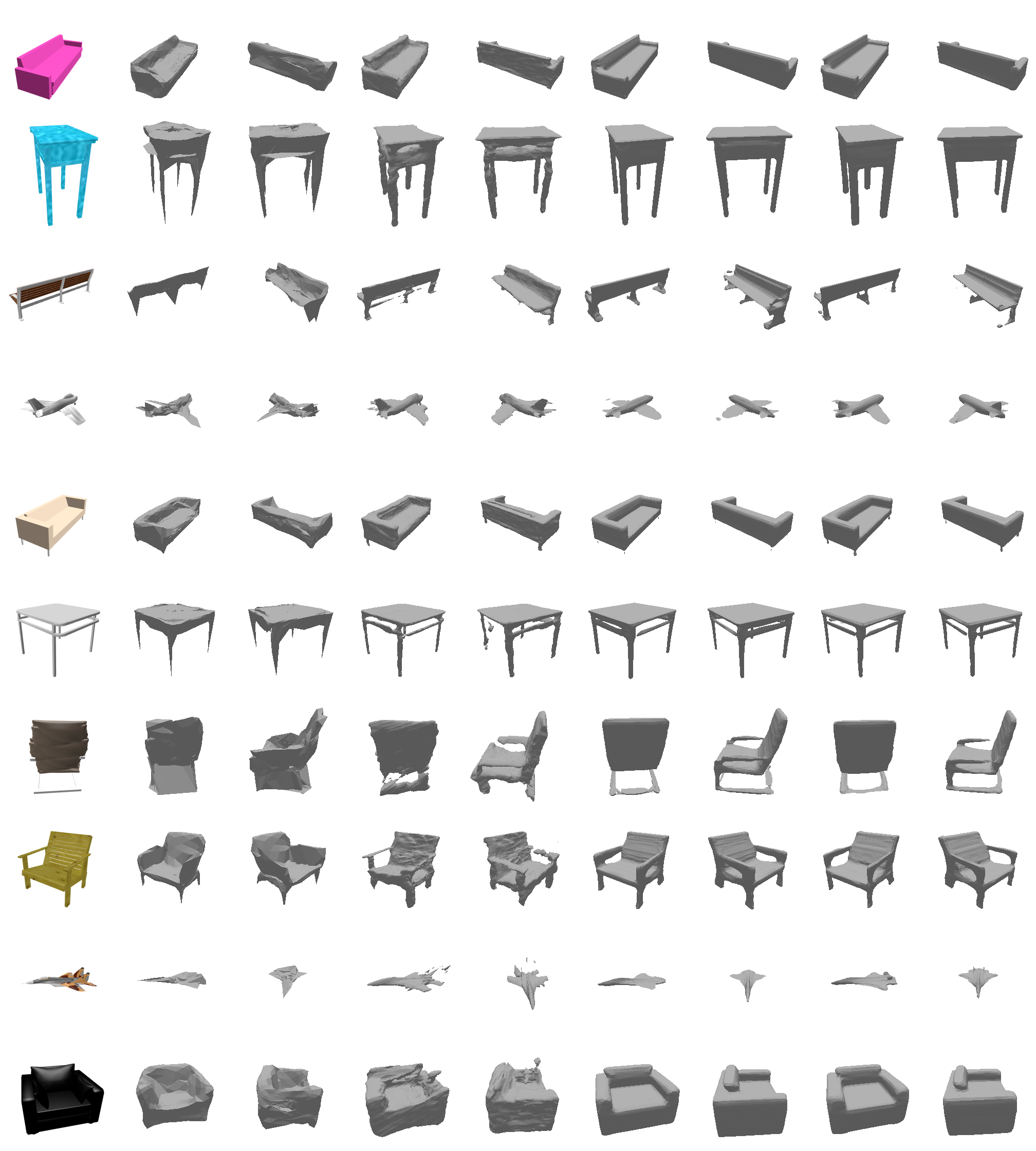}
			\put(2,101){\small{Image }}
			\put(14,101){\small{Pixel2Mesh \cite{Wang18e}}}
		    \put(36,101){\small{DISN \cite{Xu19b}}}
			\put(54,101){\small{\textit{DeepMesh}(raw)}}
			\put(75,101){\small{\textit{DeepMesh}}}
			\end{overpic}
		\end{center}
		\vspace{-10pt}
		\caption{\textbf{Comparative results for SVR on ShapeNet.} }
\label{fig:ShapeNet_qual}
\vspace{-6pt}
\end{figure*}

We provide additional qualitative comparative results for ShapeNet in Fig.~\ref{fig:ShapeNet_qual}. 
Fig \ref{fig:failure} depicts failure cases, which we take to be samples for which the refinement does not bring any improvement. These can mostly be attributed to topological errors made by the refinement process. In future work, we will therefore introduce loss functions that favor topological accuracy~\cite{Mosinska18,Oner21a}.  

\subsection{Aerodynamic Shape Optimization}

\begin{figure*}[t]
		\vspace{-3pt}
		\begin{center}
			\begin{overpic}[clip, trim=0.1cm 3.5cm 0.1cm 3cm,width= \textwidth]{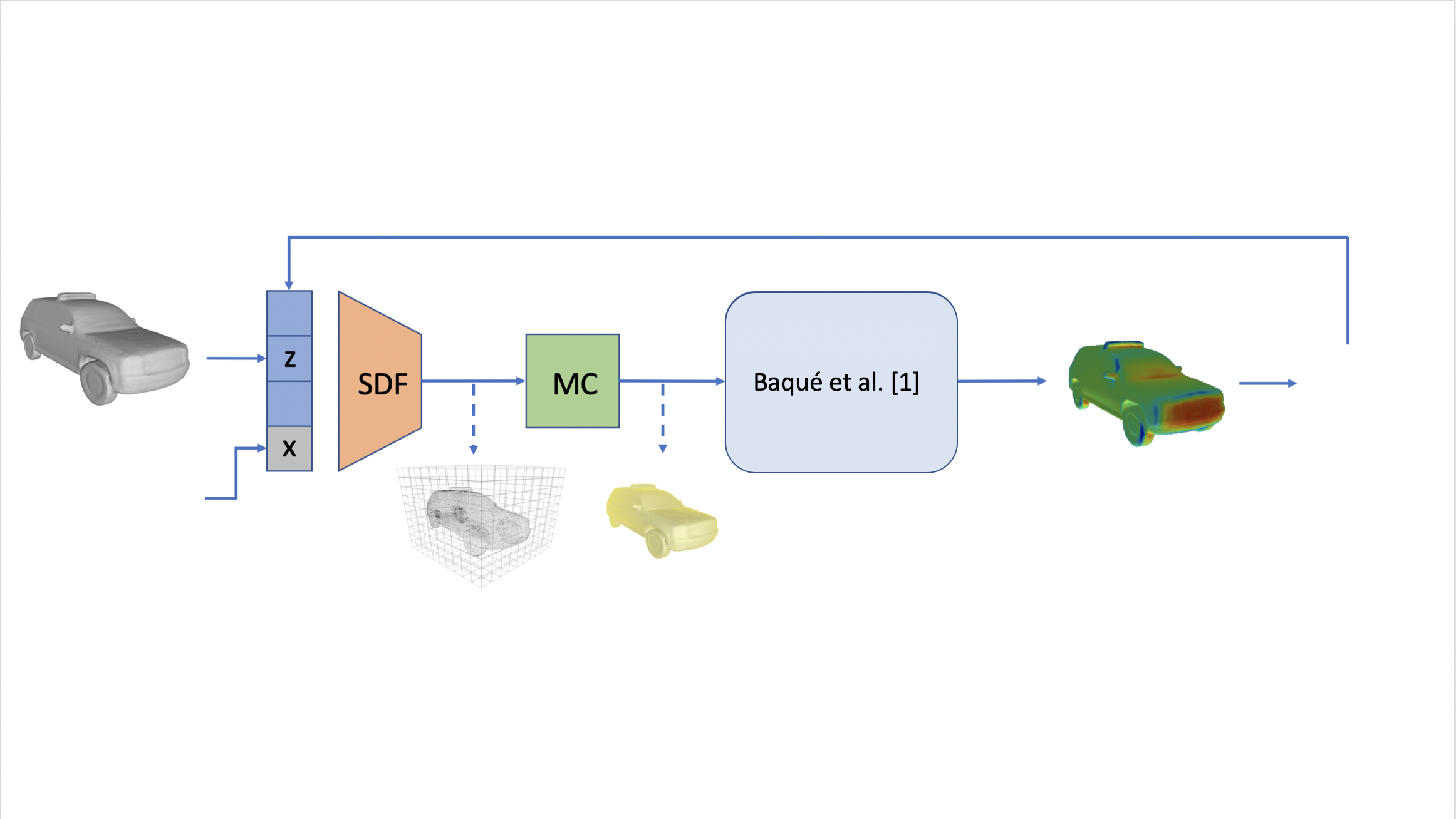}
			    \put(3,13){\small{Grid Points}}
			    
			    \put(28,4){\tiny{Predicted SDF}}
			    
			    \put(40,4){\tiny{Reconstructed Mesh}}
			    
			    \put(40,33){\small{DeepMesh Gradient (Theorem 1)}}
			    
    			\put(90,21){$\mathcal L_{\text{task}}$}
			\end{overpic}
		\end{center}
		\vspace{-6pt}
		\caption{\textbf{Aerodynamic optimization pipeline.}
		     We encode a shape we want to optimize using \textit{DeepSDF} (denoted as \textbf{SDF} block on the figure) and obtain latent code $\mathbf z$. Then we start our iterative process. First, we assemble an Euclidean grid and predict SDF values for each node of the grid. On this grid we run the Marching Cubes algorithm (\textbf{MC}) to extract a surface mesh. We then run the obtained shape through a Mesh CNN (\textbf{CFD}) to predict pressure field from which we compute drag as our objective function. Using the proposed algorithm we obtain gradients of the objective w.r.t. latent code $z$ and do an optimization step. The loop is repeated until convergence. 
		}
\label{fig:cfd_pipeline}
\end{figure*}

Here we provide more details on how we performed the aerodynamic optimization experiments presented in the main manuscript.  The overall pipeline for the optimisation process is depicted in Fig.~\ref{fig:cfd_pipeline}, and additional optimization results are shown in Fig.~\ref{fig:cfd_opta}.

\subsubsection{Dataset}

As described in the main manuscript, we consider the car split of the ShapeNet~\cite{Chang15} dataset for this experiment.  Since aerodynamic simulators typically require high quality surface triangulations to perform CFD simulations reliably, we (1) follow~\cite{Sin13} and automatically remove internal part of each mesh as well as re-triangulate surfaces and (2) manually filter out corrupted surfaces. After that, we train a DeepSDF auto-decoder on the obtained data split and, using this model, we reconstruct the whole dataset from the learned parameterization. The last step is needed so that to provide fair initial conditions for each method of the comparison in Tab. 3 of the main manuscript, that is to allow all approaches to begin optimization from identical meshes.

We obtain ground truth pressure values for each car shape with OpenFoam~\cite{Jasak07}, setting an \textit{inflow velocity} of $15$ meters per second and airflow \textit{density} equal $1.18$.
Each simulation was run for at most $5000$ time steps and took approximately 20 minutes to converge. 
Some result of the CFD simulations are depicted in the top row of Fig.~\ref{fig:cfd_prediction}.

We will make both the cleaned car split of ShapeNet and the simulated pressure values publicly available.

\subsubsection{CFD prediction}

\begin{figure}[t]
		\vspace{-3pt}
		\begin{center}
			\begin{overpic}[clip, trim=0.0cm 0cm 0cm 0cm,width= 0.4\textwidth]{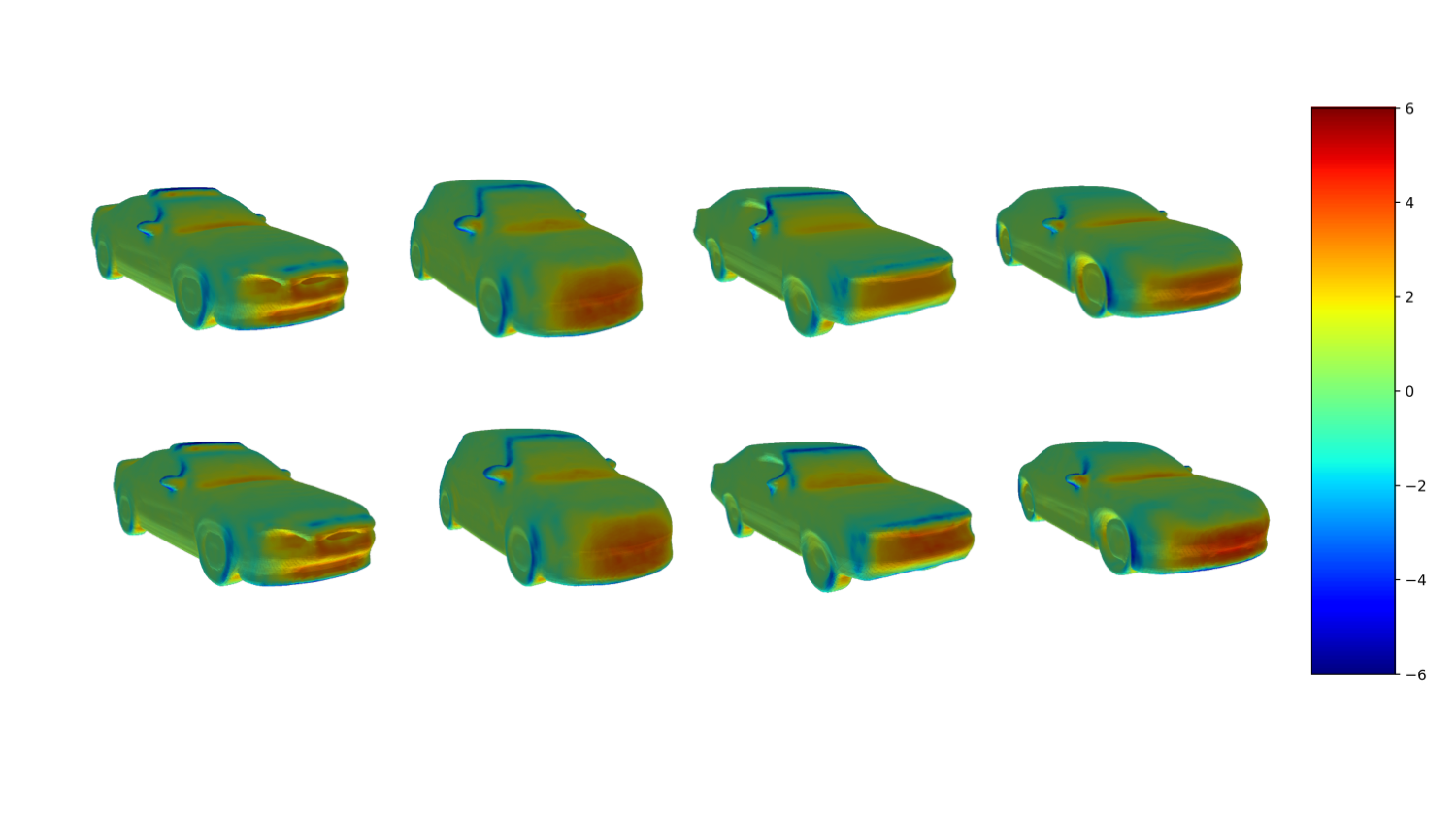}
    			\put(-2,32){\rotatebox{90}{\small{Simulated}}}
    			\put(-2,14){\rotatebox{90}{\small{Predicted}}}
			\end{overpic}
		\end{center}
		\vspace{-30pt}
		\caption{\textbf{Simulated and predicted pressure fields.}
		Pressure fields for different shape simulated with OpenFoam (top) and predicted by Convolutional Neural Network (bottom). }
\label{fig:cfd_prediction}
\end{figure}

We train a Mesh Convolutional Neural Network to regress pressure values given an input surface mesh, and then compute aerodynamic drag by integrating the regressed field. Specifically, we used the dense branch of the architecture proposed in~\cite{Baque18} and replaced Geodesic Convolutions~\cite{Monti17} by Spline ones~\cite{Fey18} for efficiency. The predicted and simulated pressure values are compared in Fig.~\ref{fig:cfd_prediction}.


\subsubsection{Implementation Details}

In this section we provide the details needed to implement the baselines parameterizations presented in the main manuscript. \comment{To foster reproducibility, we will make our code publicly available.}

\begin{itemize}
    \item\textbf{Vertex-wise optimization}
    In this baseline, we optimize surface geometry by flowing gradients directly into surface mesh vertices, that is without using a low-dimensional parameterization.
 In our experiments, we have found this strategy to produce unrealistic designs akin to adversarial attacks that, although are minimizing the drag predicted by the network, result in CFD simulations that do not convergence.
    This confirms the need of using a low-dimensional parameterization to regularize optimization.

    \item\textbf{Scaling}
    We apply a function $f_{C_x, C_y, C_z} (V) = (C_x V_x, C_y V_y, C_z V_z)^T$ to each vertex of the initial shape.
    Here $C_i$ are 3 parameters describing how to scale vertex coordinates along the corresponding axis.
    As we may see from the Tab. 3 of the main manuscript, such a simple parameterization already allows to improve our metric of interest.

    \item\textbf{FreeForm}
    Freeform deformation is a very popular class of approaches in engineering optimization.
    A variant of this parameterization was introduced in~\cite{Baque18}, where it led to good design performances.
    In our experiments we are using the parameterization described in~\cite{Baque18} with only a small modification:
    to enforce the car left and right sides to be symmetrical we square sinuses in the corresponding terms.
    We also add $l_2$-norm of the parameterization vector to the loss as a regularization.

    \item\textbf{PolyCube}
    Inspired by~\cite{Umetani18} we create a grid of control points to change the mesh.
    The grid size is $8 \times 8 \times 8$ and it is aligned to have $20\%$ width padding along each axis.
    The displacement of each control point is limited to the size of each grid cell, by applying $tanh$.
    During the optimization we shift each control point depending on the gradient it has and then tri-linearly interpolate the displacement to corresponding vertices.
    Finally, we enforce the displacement field to be regular by using Gaussian Smoothing ($\sigma = 1$, kernel size $= 3$). This results in a parameterization that allows for deformations that are very similar to the one of ~\cite{Umetani18}.
    
\end{itemize}


\begin{figure*}[t]
		\vspace{-3pt}
		\begin{center}
			\begin{overpic}[clip, trim=4cm 3.5cm 3cm 2.5cm,width= \textwidth]{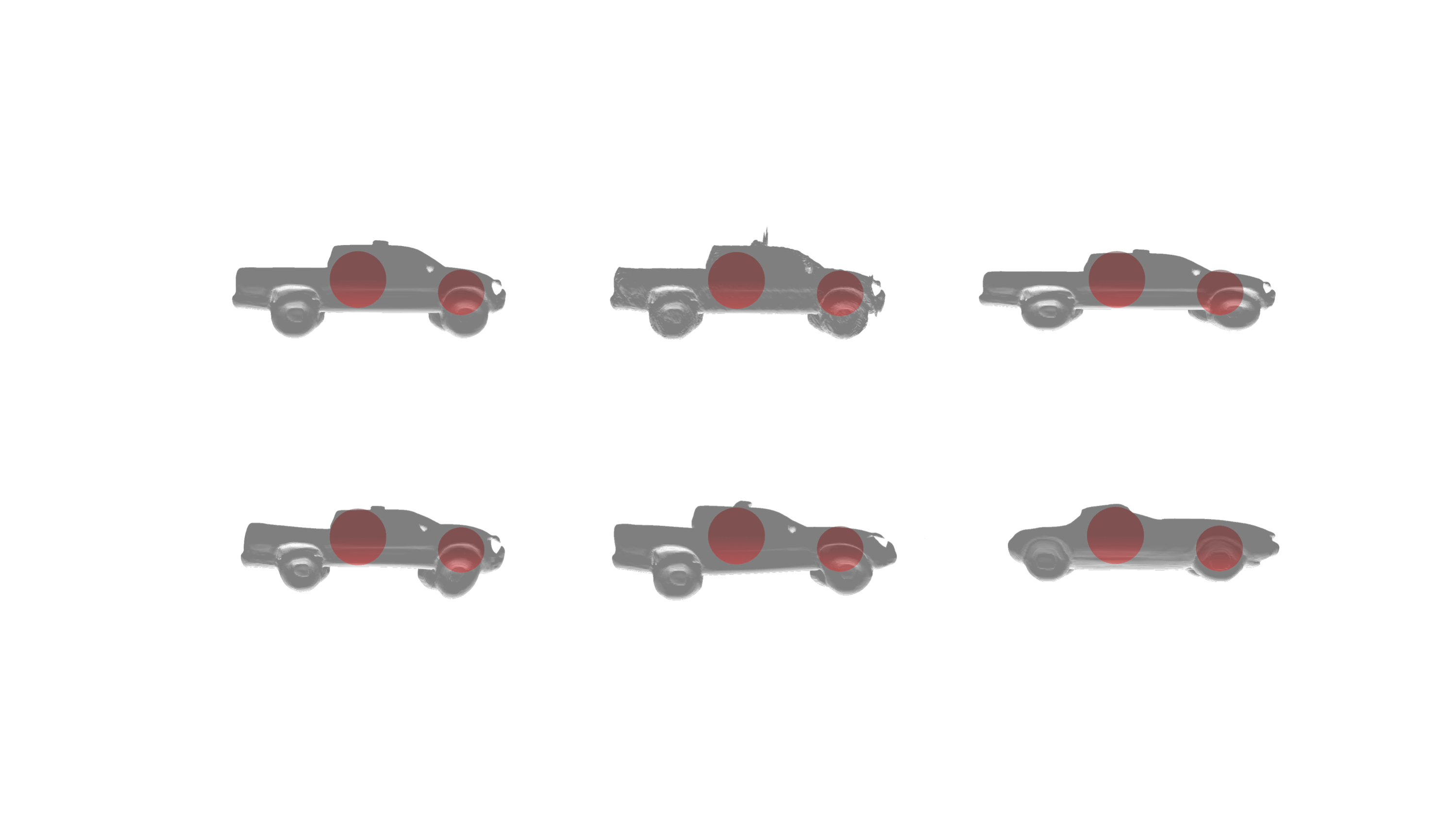}
			   \put(12,40){\small{Initial Shape}}
			   \put(45,40){\small{Vertex-vise}}
			   \put(80,40){\small{Scaling}}
			   
			   \put(13,17.5){\small{FreeForm}}
			   \put(46,17.5){\small{PolyCube}}
			   \put(80,17.5){\small{DeepMesh}}
			    
			\end{overpic}
		\end{center}
		\vspace{-6mm}
		\caption{{\textbf{Preserving space for the driver and engine.} We define a loss function $\mathcal L _{\text{constraint}}$ that forces the reconstructed shape to contain the red spheres. 
		    		    The spheres are shown overlaid on the initial shape and then on the various results. Because the constraints are soft, they can be slightly violated.}}
\label{fig:cfd_constraints}
\end{figure*}

\subsubsection{Additional Regularization for DeepMesh}

{As mentioned in the results section, to prevent the surface from collapsing to a point, we add the set of soft-constraints depicted by Fig.~\ref{fig:cfd_constraints} to preserve space for the driver and engine and define a loss term  $\mathcal L _{\text{constraint}}$. To avoid generating unrealistic designs, we also introduce an additional regularization term $\mathcal L _{\text{reg}}$ that prevents $\bz$ from straying to far away from known designs. }
We take it to be
\begin{align}
\mathcal L _{\text{reg}} = \alpha \sum_{\mathbf z' \in \mathcal{Z}_k} \frac{||\mathbf z - \mathbf z'||_2^2}{|\mathcal{Z}_k|} \; ,
\end{align}
where $\mathcal{Z}_k = {\mathbf z_0, \mathbf z_1, \ldots, \mathbf z_k}$ denote the $k$ closest latent vectors to $\mathbf z$ from the training set. In our experiments we set $k = 10$, $\alpha = 0.2$. Minimizing $\mathcal L _{\text{reg}}$  limits exploration of the latent space, thus guaranteeing more robust and realistic optimization outcomes.

In our aerodynamics optimization experiments, different initial shapes yield different final ones. We speculate that this behavior is due to the presence of local minima in the latent space of MeshSDF, even though we use the Adam optimizer~\cite{Kingma14a} , which is known for its ability to escape some of them. We are planning to address this problem more thoroughly in future.

\comment{
    Finally we end up with 
    $$\mathcal L _{\text{reg}} = \alpha \sum_{z' \in \mathcal{Z}} \frac{||z - z'||_2^2}{|\mathcal{Z}|}$$
    , where $\mathcal{Z} = {z_0, z_1, \ldots, z_k}$ are $k$ closest to $z$ latent vectors from the training set of DeepSDF.
    In our experiments $k = 10$, $\alpha = 0.2$.
    This regularization is more conservative, but stable and guarantees more realistic optimisation outcomes.
}

\begin{figure*}[t]
        \vspace{20pt}
		\begin{center}
			\begin{overpic}[clip, trim=0.0cm 0cm 0 0.0cm,width=1.0\textwidth]{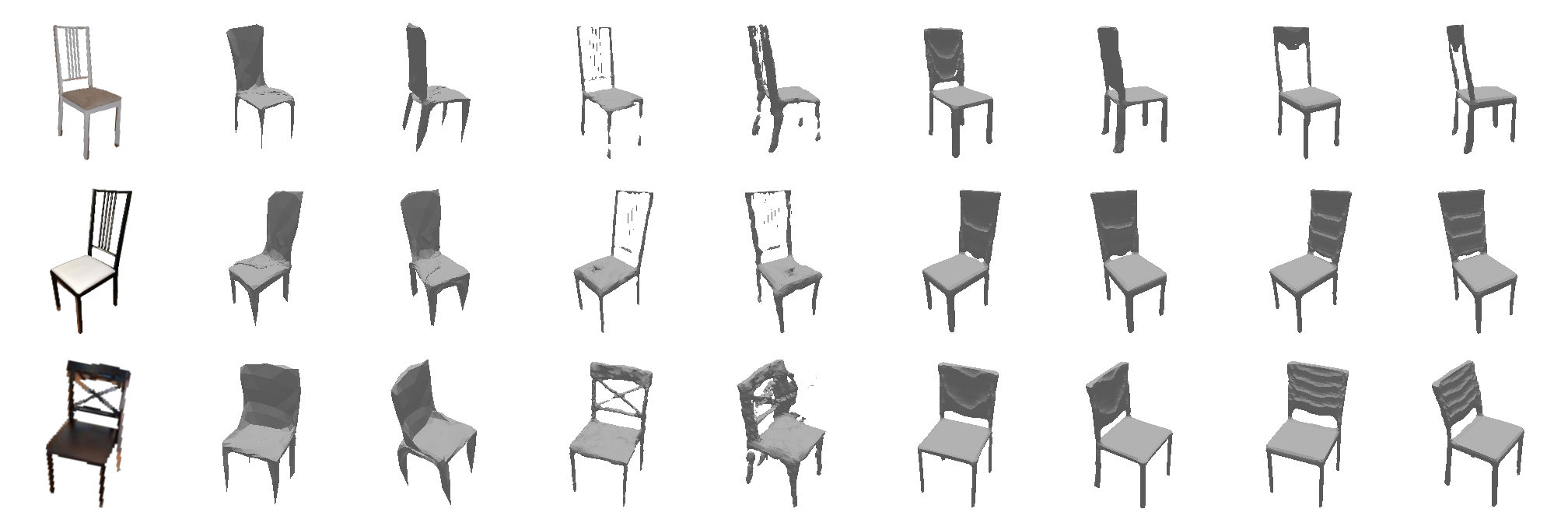}
			\put(2,35){\small{Image }}
			\put(15,35){\small{Pixel2Mesh \cite{Wang18e}}}
		    \put(39,35){\small{DISN \cite{Xu19b}}}
			\put(60,35){\small{\textit{DeepMesh}(raw)}}
			\put(85,35){\small{\textit{DeepMesh}}}
			\end{overpic}
		\end{center}
		\vspace{-10pt}
		\caption{\textbf{Failure cases for SVR on Pix3D.} Reconstruction refinement based on $L_1$ silhouette distance or {chamfer matching} fails to capture fine topological details for challenging samples.  }
\label{fig:failure}
\vspace{-10pt}
\end{figure*}
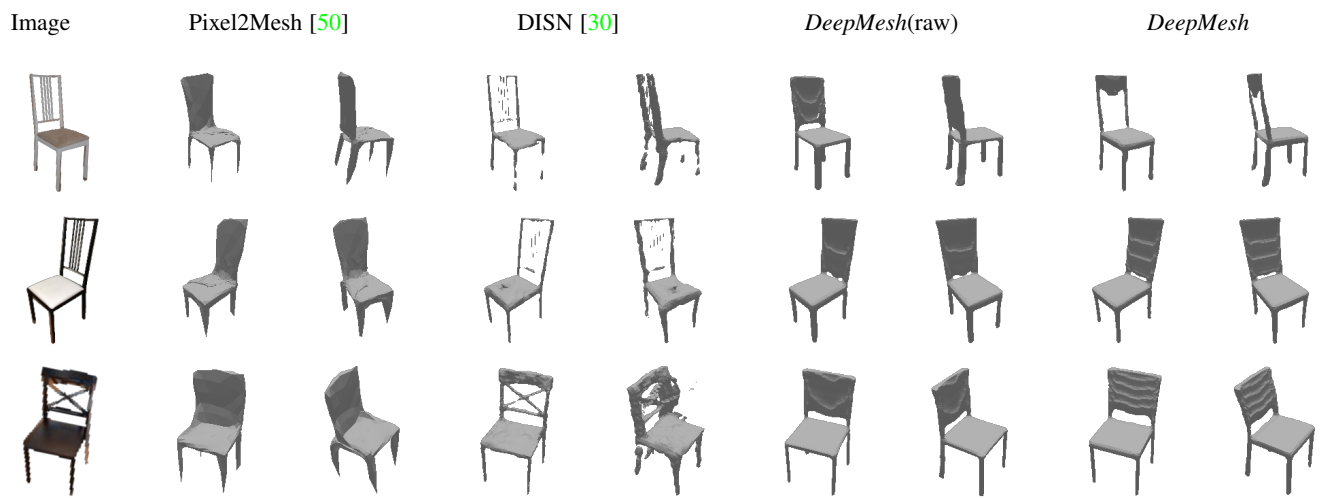
\begin{figure}[h!]
		\begin{center}
			\vspace{-5pt}
			\begin{overpic}[clip, trim=0.0cm 5.0cm 0.0cm 6.0cm,width= 1\textwidth]{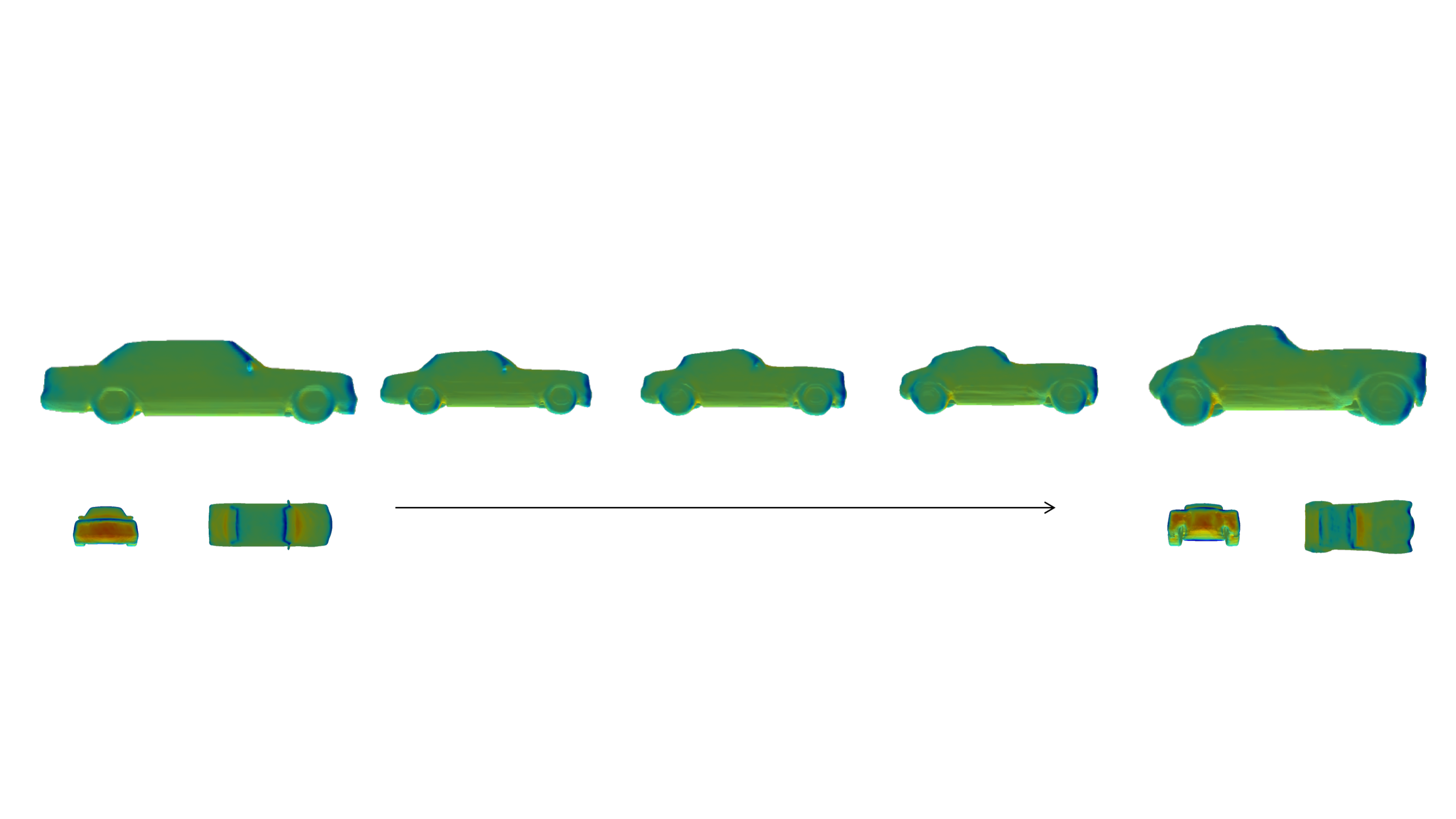}
			\end{overpic}
			\begin{overpic}[clip, trim=0.0cm 6.0cm 0.0cm 5.0cm,width= 1\textwidth]{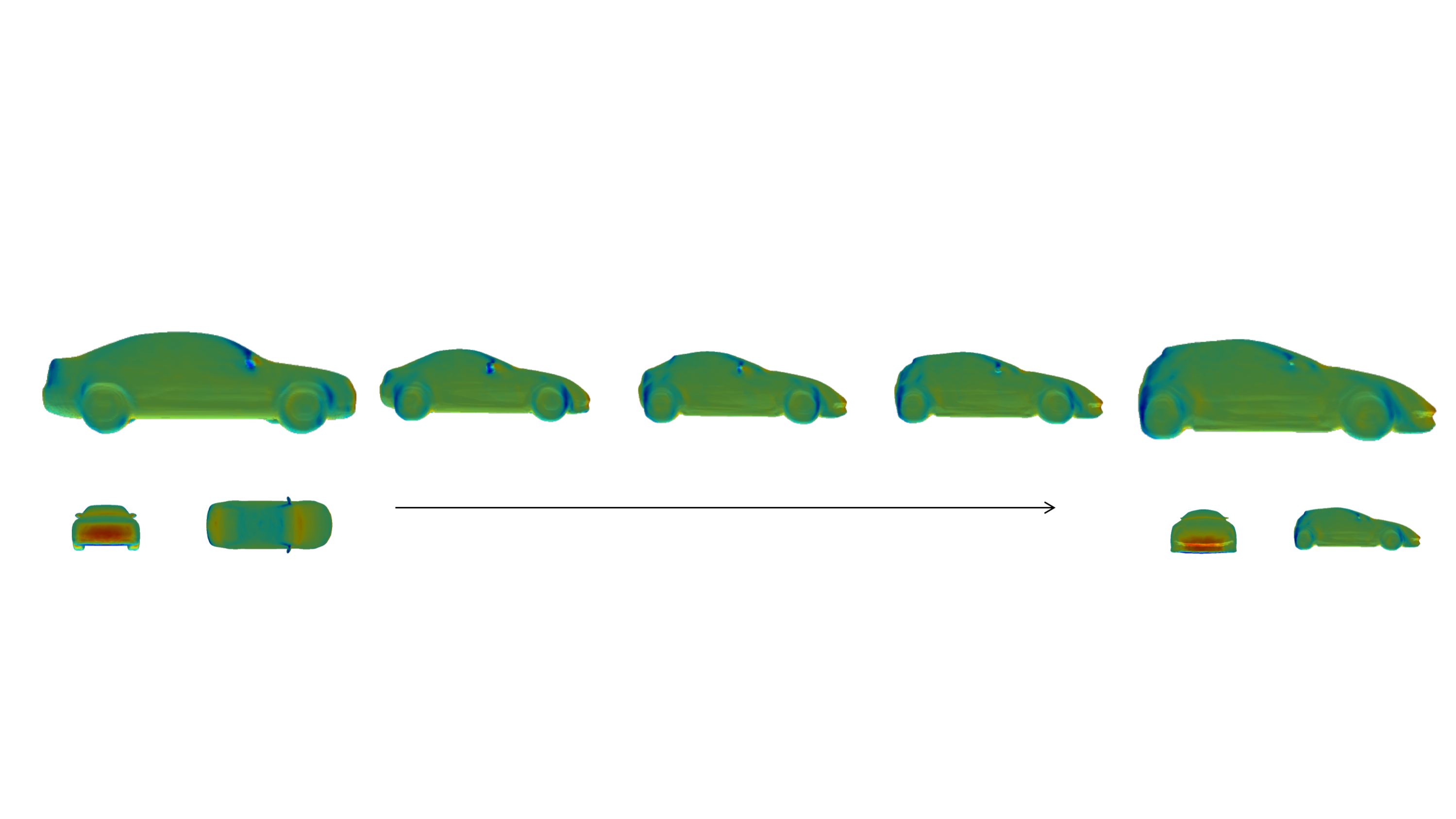}
			\end{overpic}
			\begin{overpic}[clip, trim=0.0cm 6.0cm 0.0cm 5.0cm,width= 1\textwidth]{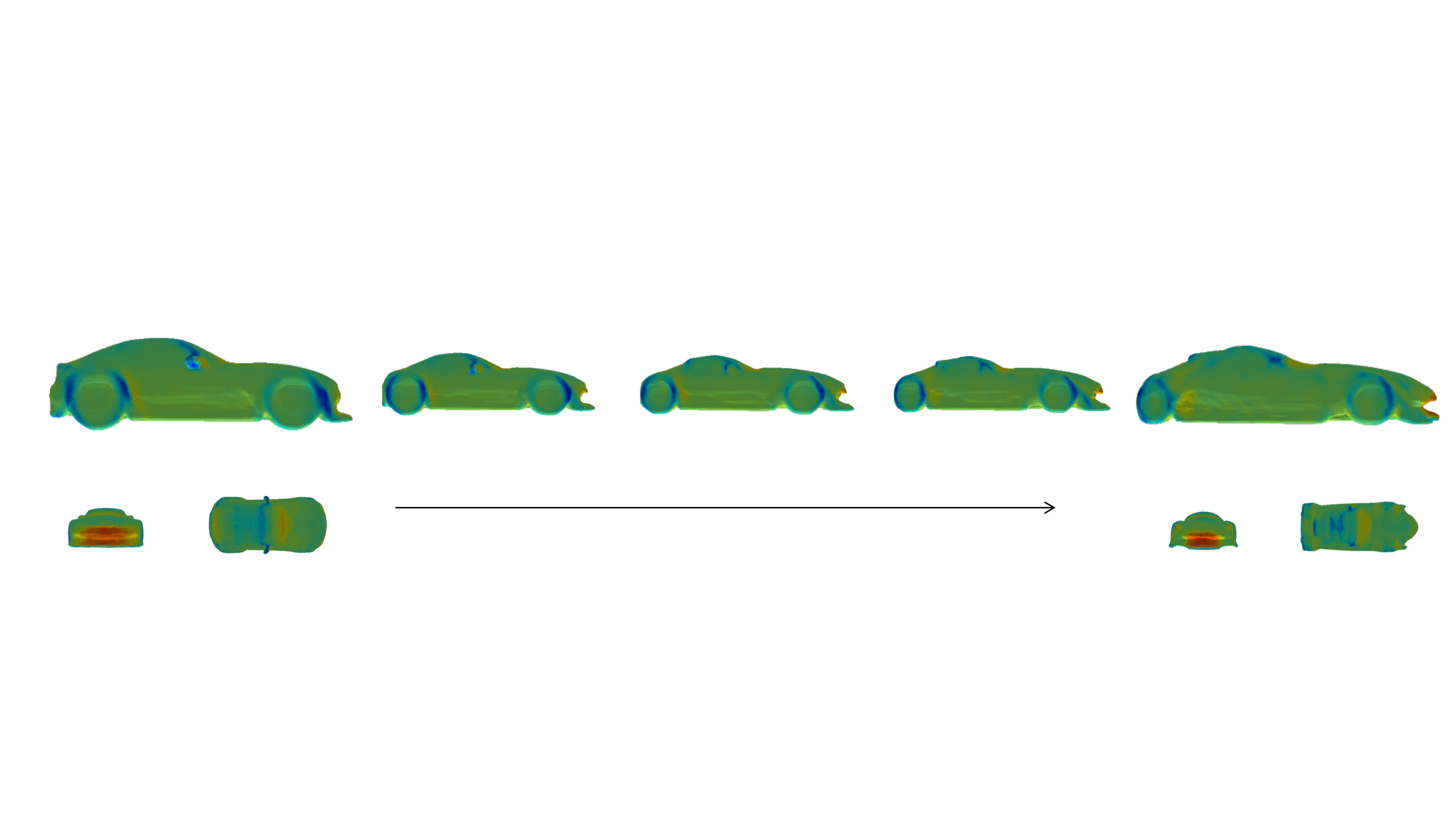}
			\end{overpic}
			\begin{overpic}[clip, trim=0.0cm 6.0cm 0.0cm 5.0cm,width= 1\textwidth]{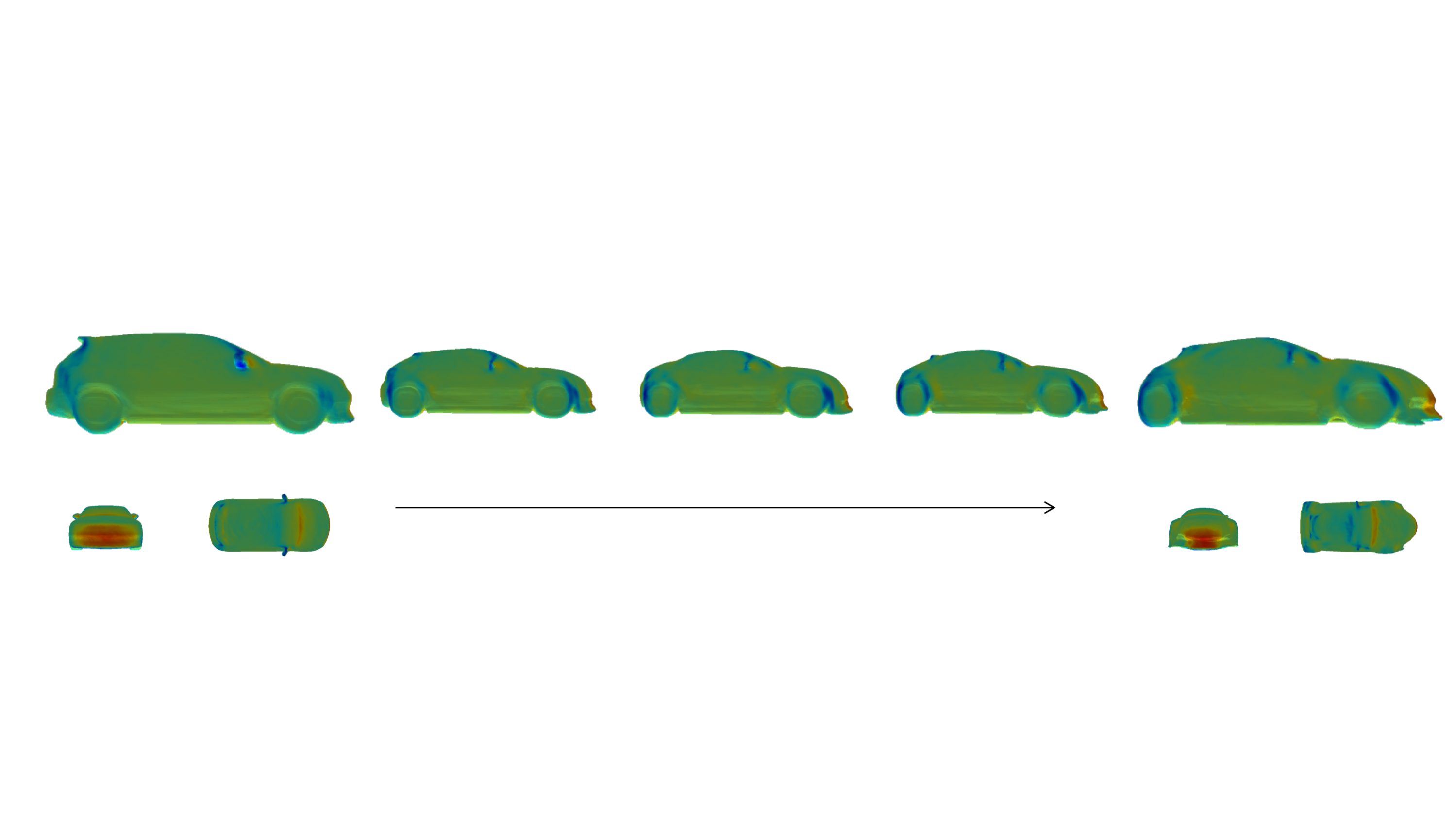}
			\end{overpic}
			\begin{overpic}[clip, trim=0.0cm 6.0cm 0.0cm 5.0cm,width= 1\textwidth]{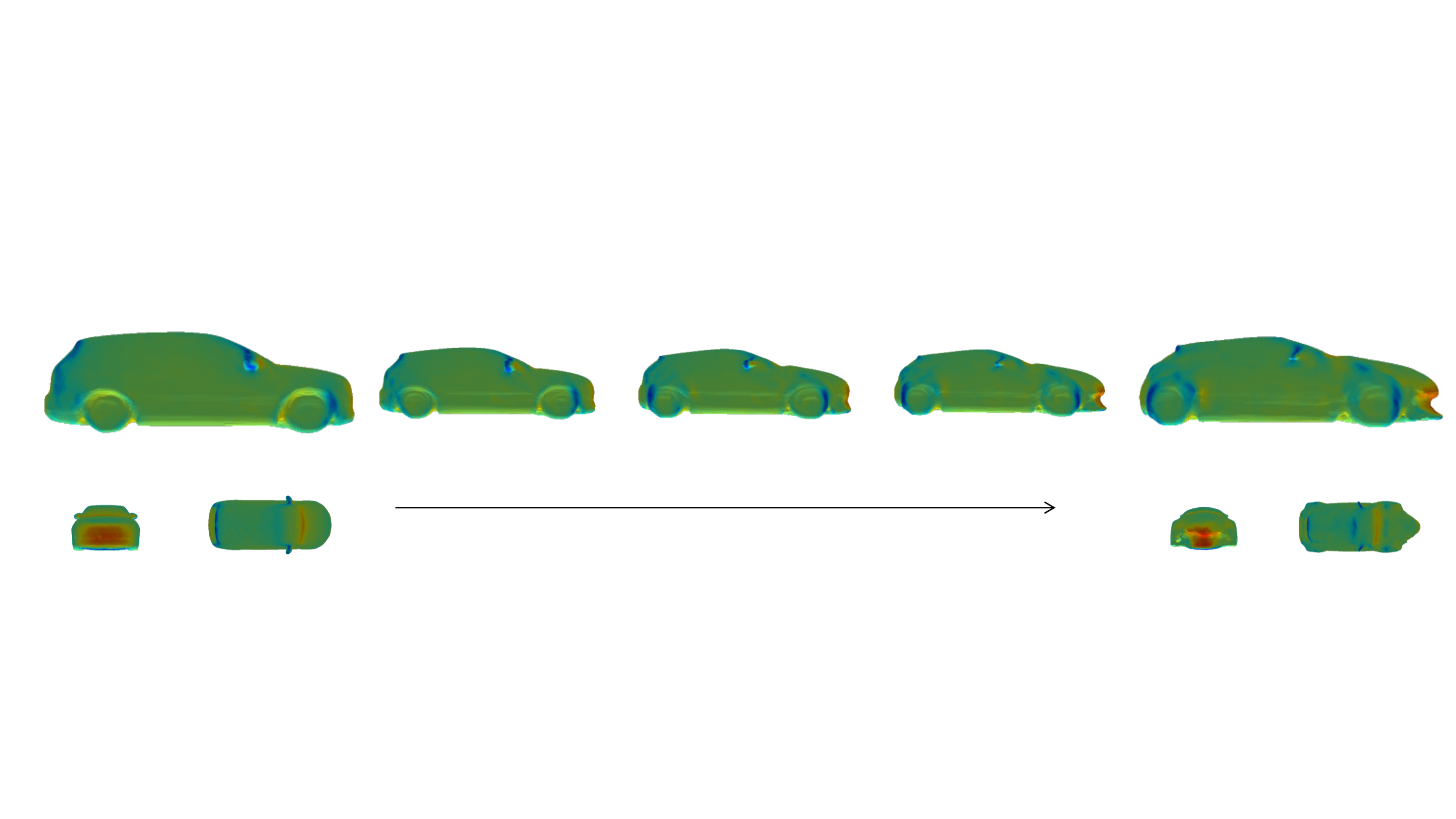}
			\end{overpic}
			\begin{overpic}[clip, trim=0.0cm 6.0cm 0.0cm 5.0cm,width= 1\textwidth]{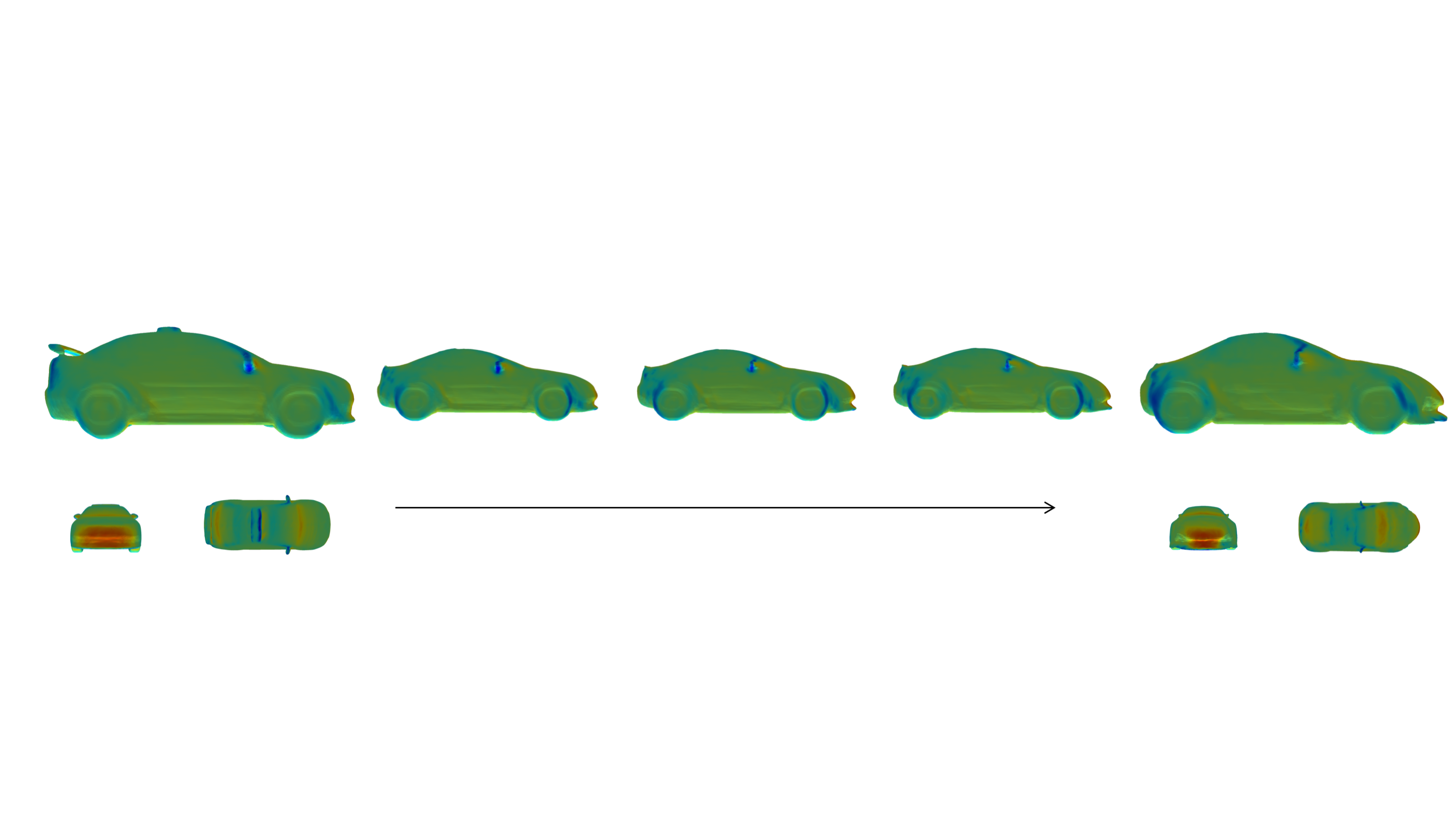}
			\end{overpic}
		\end{center}
		\caption{\textbf{DeepMesh aerodynamic optimizations.}}
\label{fig:cfd_opta}
\end{figure}










\end{document}